# Advancing Cognitive Pairwise Comparisons for Human-Centric Artificial Intelligence and Data Science: A Human-in-the-Loop Framework

Kevin Kam Fung YUEN*

School of Science, Monash University Malaysia

Jalan Lagoon Selatan, 47500 Bandar Sunway

Selangor Darul Ehsan, Malaysia

Email: kevin.yuen@monash.edu; kevinkf.yuen@gmail.com;

ORCID: https://orcid.org/0000-0003-1497-2575

**Abstract.** Human-centric artificial intelligence and data science (HCAD) systems increasingly involve human-in-the-loop (HITL) interventions to overcome the limitations of fully automated machine learning algorithms—such as automation bias and the inability to incorporate tacit domain knowledge—and continuously improve predictive performance. Data transformations, feature selection, feature weighting, and model selection usually require laborious human involvement across iterations of artificial intelligence and data science (AI&DS) pipelines, making development a challenging and time-consuming process. The Cognitive Pairwise Comparison (CPC) framework is proposed as a human-centric methodology to elicit preferences and priorities in a structured format. To establish essential facets of human-centricity, this research advances the CPC methodology so that user preference elicitation can be systematically quantified in these challenging AI&DS tasks. To demonstrate the usability and validity of these facets, this study presents several applications within movie recommender systems using the MovieLens dataset. It exemplifies how the extended CPC framework can facilitate interactions between human judgments and data across supervised learning, dimensionality reduction, and unsupervised learning algorithms.

**Keywords**: Human-in-the-Loop (HITL), Cognitive Pairwise Comparisons (CPC), Human-Centric AI (HCAI), preference elicitation, machine learning pipelines, feature and model selection, recommender systems.

## 1. Introduction

Emerging Artificial Intelligence and Data Science (AI&DS) technologies broadly disrupt human lives. They reshape how innovation is organized, overcome human constraints in information processing, assist in generating breakthroughs, and exert a transformative impact on the digital structure of organizations (Haefner et al., 2021). Nevertheless, the ubiquity of AI&DS creates significant challenges, largely stemming from the marginalization of human interaction. Everyday phenomena—what we read and watch, how we connect with others and build relationships, and how decisions regarding employment, loans, and insurance are made—are increasingly influenced by mathematical models and their underlying data-driven systems (Thurman et al., 2019). Consequently, AI capabilities are often overhyped; without grounded human input and domain understanding, AI agents act merely as parrots that mimic human language without genuine comprehension. To bridge this fundamental gap, technical development must be anchored within broader human contexts. Highlighting this necessity, Carter et al. (2025) examine the practical mechanics of cross-disciplinary collaboration between AI and Humanities, Social, and Economic Sciences (HSES) experts. Based on an 18-month initiative, they conclude that integrating these fields for responsible AI development demands dedicated resources, a long-term vision, and room for continuous experimentation.

AI "thinking" is fundamentally a bundle of algorithms created by highly specialized human designers. AI cannot genuinely think or understand; rather, human users justify that AI is "intelligent" only when the algorithms output sound, actionable responses. For example, actionable insights or recommendation choices are not consumed by machines, but by human decision-makers. Because the creation of AI&DS applications inherently requires human intelligence, diverse development teams build AI products based on their own specific selections of platforms, rules, packages, and datasets. In other words, AI&DS products created by humans are seldom perfect and frequently make mistakes—often failing to align with designers' initial assumptions and operational definitions. Consequently, there are diverse popular alternatives among AutoML platforms, such as such as Google Cloud AutoML, currently offered as part of the Google Gemini platform (Google, 2026), Amazon SageMaker AutoPilot (Amazon Web Services, 2026), IBM AutoAI (IBM, 2026), Microsoft Automated ML (Microsoft, 2026), Auto-WEKA (Kotthoff et al., 2019; Thornton et al., 2013), Databricks AutoML (Databricks, 2026), SAS Model Studio (SAS, 2026), and H2O Driverless AI (H2O.ai, 2026). When these platforms are used to solve the exact same problem, they frequently yield different performance results and error profiles, because the underlying automation algorithms reflect the distinct initial assumptions of their human designers. Thus, human involvement remains essential in the decision-making process.

Broussard (2018) used the term "Artificial Unintelligence" to describe how computer algorithms can fundamentally misunderstand the world. Because people often assume computers are more objective than humans, they frequently attribute algorithmic mistakes to human user error rather than questioning the logic within thousands of lines of code or the biases in the sample data. Experiments by Dietvorst et al. (2015) on understanding human confidence in algorithms concluded that people may choose an algorithm if they have not yet seen its results (even if those results surpass their own). However, once people observe an algorithm making mistakes, they strongly avoid using it; conversely, if they can modify the algorithm even slightly after observing its performance, their willingness to adopt it increases significantly. Therefore, combining human judgment with algorithmic processing helps users feel more confident. Building on this, Korowicz and Thommes (2025) evaluated the impact of algorithmic, expert, and hybrid feature selection methods on AI-supported decision-making, revealing that while users strongly prefer a combined human-algorithmic development approach, neither their baseline preferences nor having model-selection choice significantly alters their actual reliance on AI advice. Ultimately, establishing a human-in-the-loop (HITL) framework helps guard against the detrimental effects of erroneous algorithmic recommendations (De-Arteaga et al., 2020).

While defining the exact nexus among machine learning (ML), artificial intelligence (AI), and data science (DS) is beyond the scope of this study, these fields are deeply interconnected. ML serves as an essential technique underpinning AI, and AutoML represents a subset of AutoAI, which overlaps heavily with AutoDS. Because these domains collectively rely on "learning from data," this study adopts the unified term AI&DS.

Despite the recognized conceptual need for HITL frameworks, a critical operational gap remains. While current literature agrees that human expertise must be integrated into AI pipelines, existing methods largely rely on qualitative design guidelines, post-hoc interpretability tools, or simplistic behavioral feedback loops. They lack a rigorous, interval-scale mathematical mechanism capable of translating tacit human judgments into precise numerical weights that can be programmed directly into low-level pipeline operations—such as data transformation, feature selection, and model tuning.

Cognitive Pairwise Comparisons (CPC) is a mathematically rigorous elicitation method that captures tacit human preferences through structured pairwise evaluations and translates them into explicit, quantitative paired interval-scale weights (Yuen, 2009, 2012). To bridge the specific operational gap identified above, this research extends this baseline CPC method to systematically integrate human knowledge into AI&DS algorithms, forming a sophisticated quantitative approach for solving complex decision problems. The core novelties and main technical contributions of this work are threefold:

- Mathematical Elicitation for Machine Learning Pipelines: Unlike existing tools that rely strictly on data-driven heuristics or qualitative interviews, this framework provides quantifiable formulas (Eqs. 1–7) to systematically encode human mental models into low-level pipeline operations.
- Algorithmic Integration Across Pipeline Stages: Cognitive Pairwise Comparisons (CPC) are extended beyond standard multi-criteria decision-making into programmatic algorithms, including Feature Weighting (Eq. 8), Categorical Encoding (Eqs. 9–11), Discretization (Eqs. 12–14), data splitting (Algorithm 1), forward feature selection (Algorithm 2), model selection (Eqs. 18–20), weighted distances (Eqs. 21–24), and user-preference-weighted PCA (Algorithm 3).
- Cross-Paradigm Applicability: A single quantitative preference elicitation technique is demonstrated to apply seamlessly across supervised learning, unsupervised learning, data reduction, and rule mining workflows.

The remainder of this paper is organized as follows. Section 2 reviews recent literature on Human-Centric AI&DS and Human-in-the-Loop (HITL) paradigms to highlight current methodological gaps and formally motivate this study. Section 3 introduces the proposed CPC-HCAD framework, detailing its conceptual architecture. Section 4 presents the core mathematical foundations of Cognitive Pairwise Comparisons (CPC). Building upon these foundations, Section 5 proposes specific CPC extensions tailored to embed user preferences into key HITL pipeline activities. Finally, Section 6 demonstrates the practical applicability of these proposed algorithms through a case study on a popular movie recommender dataset, followed by concluding remarks in Section 7.

## 2. Related Work and Motivations

### 2.1 Limitations of Auto AI&DS

The rise of AI&DS applications is transforming existing business processes by automating tasks, augmenting human capabilities, and replacing certain human roles, thereby reshaping educational, social, and business structures. Recent techniques in automating AI&DS creation processes—such as AutoAI or AutoML mentioned in the preceding section—aim to automate data science tasks, suggesting that an increasing portion of data science responsibilities could potentially be automated. For example, AutoDS is an AutoML system designed to leverage state-of-the-art ML automation techniques to support DS projects (D. Wang et al., 2021). Typically, users need only upload their datasets before systems like AutoDS automatically configure algorithmic hyperparameters, preprocess data, select optimal algorithm combinations, and subsequently train and deploy the model.

However, a growing body of research explores the human perspective within Auto AI&DS. The fully automated settings in which current AutoML systems operate are rarely a one-size-fits-all solution across diverse users and problem domains (Lee et al., 2020). AutoML systems still rely on considerable human effort to be effective, and even as the underlying technology improves, human oversight remains essential to ensure safety and efficacy (Crisan & Fiore-Gartland, 2021). Consequently, AutoAI systems should be designed to augment, rather than entirely automate, the complex practices of heterogeneous data science teams (D. Wang et al., 2019). While models produced by automated systems often exhibit fewer technical errors, they paradoxically yield lower human confidence scores (D. Wang et al., 2021). Increasing transparency by providing users with clearer insights into AutoML tools significantly improves user trust and understandability (Drozdal et al., 2020). Furthermore, automation bias can lead humans to blindly follow machine recommendations, failing to utilize alternative information sources or independent judgment to override erroneous output (De-Arteaga et al., 2020). To combat this, Li and Fan (2026) demonstrated that actively incentivizing human decision-makers to evaluate and correct algorithmic predictions significantly reduces bias without compromising accuracy, effectively highlighting the indispensable role of human-algorithm collaboration.

### 2.2. Human-Centric (or Centered) AI&DS

The realization of fully autonomous AI&DS remains challenging. Because the high complexity of AI&DS development demands intensive human involvement, formulating a generalized, fully automated solution is difficult; instead, Human-Centric AI&DS (HC-AI&DS) provides targeted solutions on a case-by-case basis. For example, an evaluation of commercial data preparation tools by Hameed and Naumann (2020) revealed that core tasks—such as data discovery, validation, structuring, enrichment, filtering, and cleaning—still require extensive manual intervention.

Kacprzyk et al. (2019) defined human-centric (or human-centered) computing as developing problem solutions that explicitly incorporate the human perspective—employing tools and techniques for problem formulation, solution design, and result interpretation that align with human cognitive abilities and communication skills. Advancing this paradigm, Natarajan et al. (2025) highlighted the critical distinction between "human-in-the-loop" and "AI-in-the-loop" paradigms, arguing that properly recognizing human experts as primary decision-makers is essential for the effective design and evaluation of collaborative systems.

Although fully automated machine learning is appropriate for many applications, human users often possess domain knowledge that supplements and constrains available data and analytical solutions (Gil et al., 2019). To leverage this expertise, Gil et al. (2019) proposed human-guided machine learning (HGML) as a hybrid approach that enables users to apply domain knowledge in guiding AutoML systems during algorithm selection and multi-step solution synthesis. Similarly, Lee et al. (2020) outlined a vision for a mixed-initiative machine learning environment (MILE) by rethinking the complementary roles that automation and human supervision play across the ML development lifecycle. To ensure these complex human-machine collaborations operate safely, Abeywickrama and Ramchurn (2024) proposed a formal model-checking methodology to verify, certify, and explain responsible decision-making behaviors—specifically addressing safety, controllability, and ethics within human-agent collectives. Practical implementations, such as Pandora (Nushi et al., 2018), further leverage both human- and system-generated observations to summarize conditions of system malfunction with respect to input content and system architecture.

### 2.3 Human-in-the-Loop in AI&DS

Data science projects are inherently collaborative, requiring teams to engage with diverse stakeholders and tools throughout the development lifecycle (Zhang et al., 2020). The design of interactive machine learning (ML) systems must involve users across all stages—from exploratory phases that reveal human interaction patterns to refinement stages that fine-tune interface details and evaluate alternatives (Amershi et al., 2014). Furthermore, humans actively intervene in data pipelines by discovering, capturing, curating, designing, and creating datasets (Muller et al., 2019). Gil et al. (2019) highlighted several major tasks required of human guides, including variable selection, variable requirements, instance selection, model selection, parameter settings, solution design, model assessment, model comparison, and parameter comparison. While automated systems handle key wrangling decisions on matching, mapping generation, mapping selection, format transformation, and data repair, users must provide the target data schema, partial representative instances aligned with the target, prioritization criteria for populating the target, and feedback on candidate results (Konstantinou et al., 2019). This human involvement is especially crucial during data preparation, which utilizes sequential pipeline steps to transform datasets (Hameed & Naumann, 2020) and consumes approximately 80% of data teams' workflow (Hellerstein et al., 2018; Patil, 2012). To optimize this phase, Rezig et al. (2019) presented a human-centric data-cleaning framework where users interactively detect, repair, and validate data errors while defining structural specifications.

A persistent challenge in these workflows is the disconnect between data scientists and domain experts, as both groups perceive modeling and business domains differently (Viaene, 2013). Kaur et al. (2020) identified a misalignment between data scientists' understanding of interpretability tools and their intended usage, emphasizing that overcoming this gap requires expertise in the mathematics underlying ML models and in effectively communicating information to users (e.g., through tutorials, visualizations, or interactive tools). Similarly, Hong et al. (2020) conducted 22 semi-structured interviews with industry practitioners and demonstrated that ML interpretability is contextually dependent on practitioners' organizational roles and socially negotiated through stakeholder interactions across development pipeline stages.

To synthesize these varied human interactions across the pipeline, Mosqueira-Rey et al. (2023) systematically reviewed Human-in-the-Loop (HITL) techniques, categorizing them by the degree of human agency into active learning, interactive ML, and machine teaching. Building on this concept of agency allocation, Borchers et al. (2026) introduced the Agency Allocation Framework to systematically analyze how decision-making authority is distributed between human stakeholders and automated systems, moving away from simple behavioral proxies. Punzi et al. (2026) provided a structured taxonomy of collaborative frameworks focused on uncertainty-driven abstention mechanisms, which dynamically determine when an AI should act autonomously versus when it must defer to human expertise. However, while these state-of-the-art frameworks successfully delineate qualitative control structures and theoretical agency models, existing HITL architectures still lack an interval-scale mathematical mechanism based on pairwise comparisons to programmatically embed human preferences into low-level pipeline operations—a gap directly addressed by the proposed CPC-HCAD approach.

### 2.4 Soft Computing for Human Role in AI&DS

Soft computing is an interdisciplinary field focused on designing intelligent systems to process uncertain, imprecise, and incomplete information. Human-centricity is an inherent property of how people perceive a system, process, machine, or model, while information granularity reflects human uncertainty, playing a pivotal role in soft computing (Song & Wang, 2015). Kou et al. (2020) compared five multiple-criteria decision-making (MCDM) methods to evaluate and select feature-selection techniques for text classification. Similarly, Chen et al. (2008) applied the fuzzy Analytic Hierarchy Process (AHP) to multidimensional scaling (MDS) to identify preference similarities among alternatives.

Cognitive Pairwise Comparisons (CPC) (Yuen, 2009, 2012), a soft computing technique, serves as the foundational core of the Cognitive Network Process (CNP), outperforming AHP because CPC offers a structured elicitation methodology to encode tacit human preferences into an explicit format. Applications of CPC are widely demonstrated across diverse decision domains (Goldani & Ishizaka, 2025; Goldani et al., 2023; Paucar et al., 2019, 2020; Qi et al., 2017; Ri et al., 2024; C.-H. Wang, 2018, 2019; Yuen, 2014a, 2014b, 2021). To address these limitations, this study introduces key initiatives on how CPC can be systematically utilized for HCAD, as detailed in the subsequent sections.

### 2.5 Motivations

To explicitly position the proposed CPC-HCAD framework against existing paradigms, Table 1 summarizes the core methodological distinctions among standard automated machine learning (AutoML), qualitative Human-Centric AI (HCAI) studies, and the proposed quantitative framework.

**Table 1.** Key distinctions between the proposed CPC-HCAD framework and conventional AI development paradigms

| Dimension | AutoML | Existing HCAI Studies | Proposed CPC-HCAD Framework |
|---|---|---|---|
| Human Role | Omitted / Fully automated | High-level guidance / Qualitative feedback | Quantitative preference elicitation via CPC |
| Data Transformation | Data-driven / Algorithmic | Manual or tool-assisted | Mathematically encoded human preferences (POM) |
| Pipeline Scope | Hyperparameter tuning & model selection | Task-specific interactions | Scope of pipeline shown in Figure 2 (Encoding, Splitting, Selection, PCA, Model Selection) |
| Consistency Control | None | Implicit | Accordance Index (AI) validation |

## 3. Framework of Human-Centric AI&DS with CPC

### 3.1. Hybrid Human-AI decision-making framework

The concept of Human-Centric AI and Data Science (HCAD) within this research is illustrated in Figure 1. Decision-making is typically achieved in one of three ways: utilizing human opinions through mental models, analyzing data via AI&DS algorithms, or employing a hybrid system that combines both approaches. Driven by the rapid expansion of data volume and variety, numerous statistical and machine learning algorithms have been adopted to analyze datasets of all scales, aiming to extract meaningful patterns that inform sound decision-making.

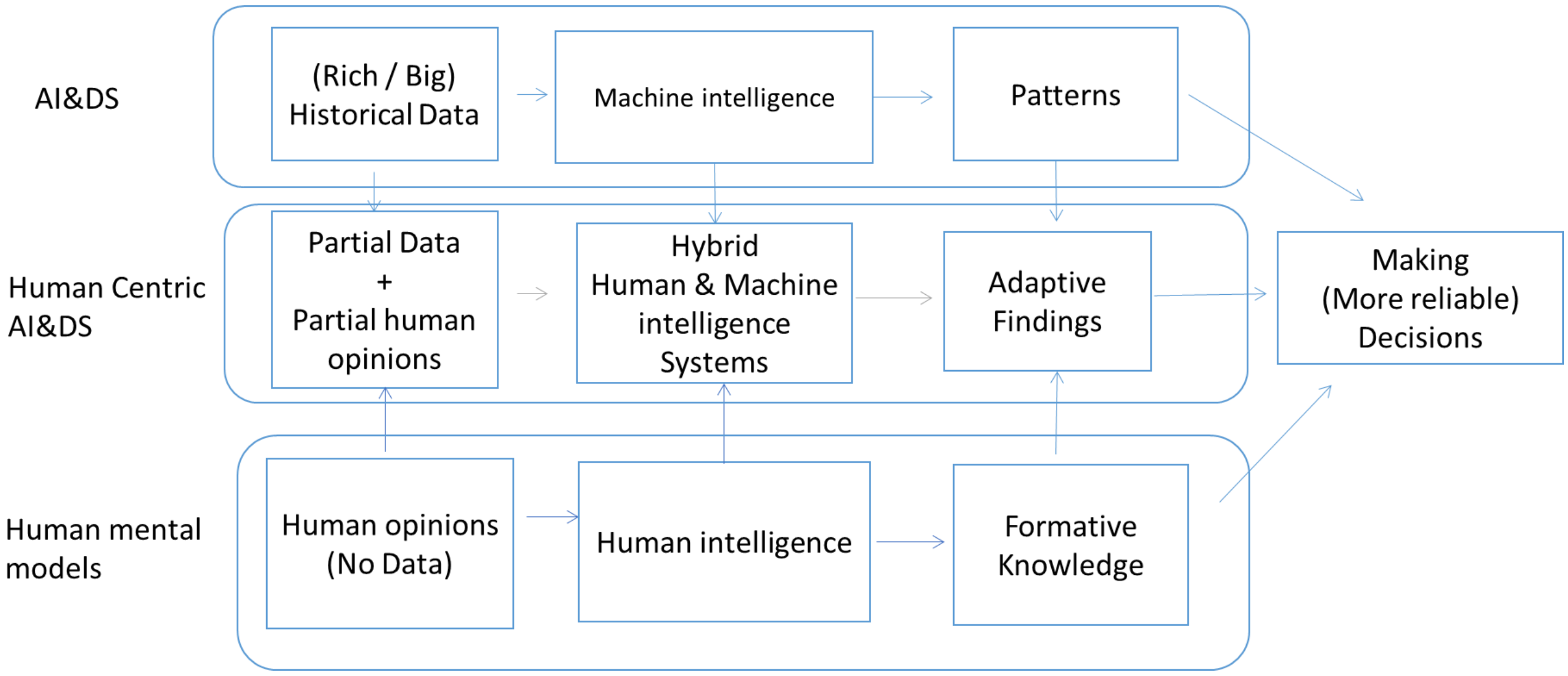

Figure 1: Hybrid Human-AI decision-making

Nevertheless, many decision problems lack historical data or reference cases; even when relevant data exists, it is often difficult or expensive to obtain. Human mental models formulate decision-making frameworks by utilizing human knowledge, judgments, preferences, and opinions as foundational inputs to construct primary solutions. Research in multi-criteria decision-making (MCDM) is closely aligned with these human mental models. In many scenarios, complex decision problems require the simultaneous analysis of partial data and the integration of human knowledge. This necessity has driven research into HCAD—such as Hybrid Human & Machine Systems (HHMS)—which integrates human expertise with data analytics algorithms to offer promising solutions for complex problems. Within the HCAD framework, human participants may include data science professionals, domain experts, or end users.

### 3.2 CPC in HCAD

Unlike prior studies that treat Cognitive Pairwise Comparisons (CPC) solely as a multi-criteria decision-making tool, this work introduces seven novel formulations that adapt CPC into programmatic data preprocessing and machine learning algorithms. The complementary strengths of humans and AI can offset each other's limitations to achieve more reliable and trustworthy decision outcomes. A primary research challenge in HCAD is determining how humans can effectively participate in conventional data science pipelines. Figure 2 presents a framework illustrating how the proposed CPC extensions can be integrated into the essential components of standard data science pipelines, as follows:

- Data preprocessing and preparation: discretization, feature selection, feature weighting, and categorical encoding.
- Model selection and evaluation: splitting datasets into subsets, such as training, validation, and testing sets.
- Data transformation: weighted distances and data reduction.

Finally, the preprocessed and transformed data can be utilized for supervised learning (regression and classification), unsupervised learning (clustering and association learning), and data reduction. Detailed mathematical formulations for these integrations are presented in the subsequent sections.

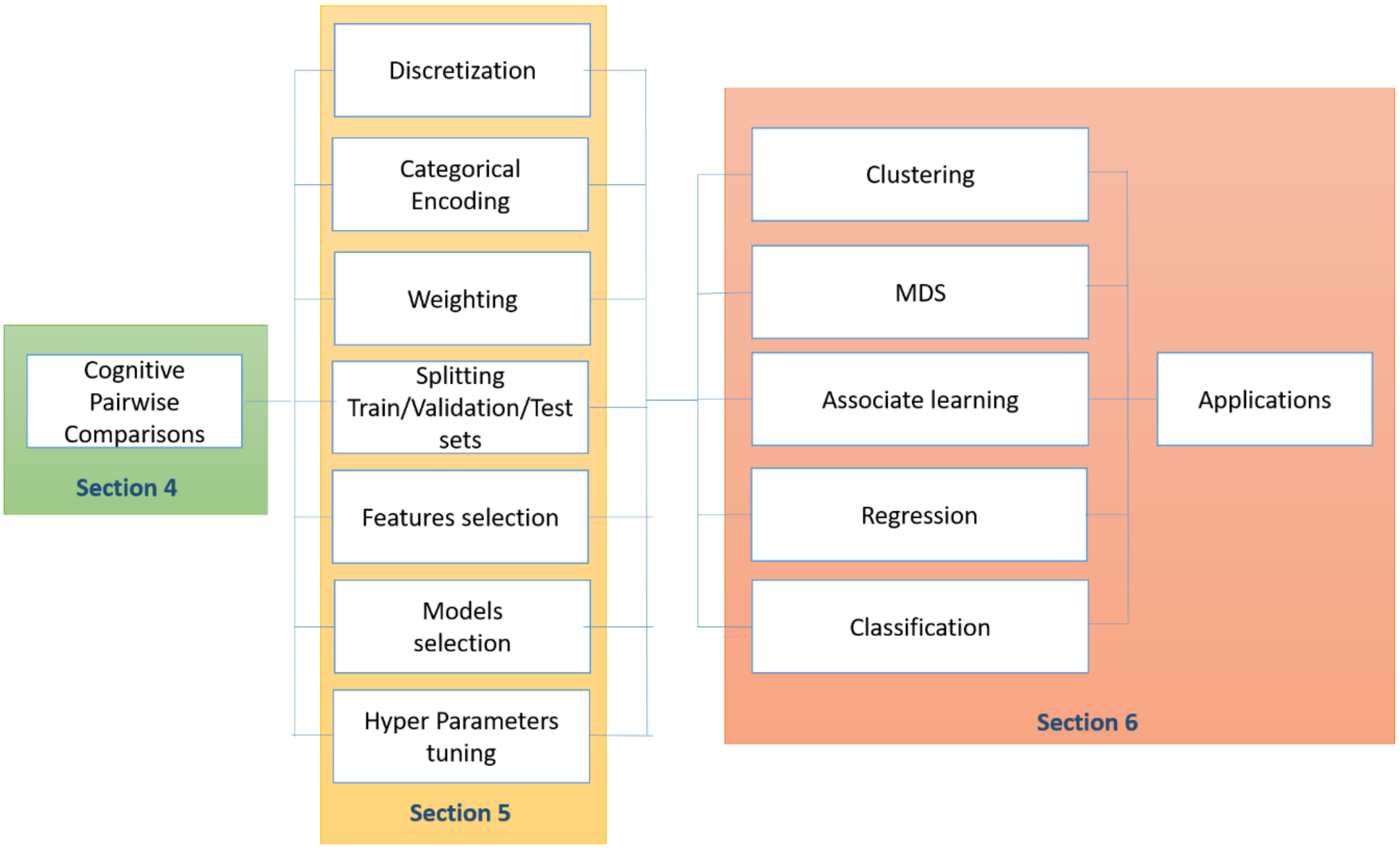

Figure 2. Cognitive Pairwise Comparisons in the Human-in-the-Loop Framework for HCAD

## 4. Cognitive Pairwise Comparisons

This study advances Cognitive Pairwise Comparisons (CPC) by extending its application as a human-centric methodology within AI&DS pipelines, as illustrated in Figure 2. The CPC procedure elicits the individual values of measured objects by constructing a Pairwise Opposite Matrix (POM). This matrix comprises comparison scores derived from evaluating all possible pairs of objects, typically collected via a survey form (an example is provided in Figure 3). The underlying mathematical formulations are detailed below.

Depending on the specific measurement scenario, a set of measured object values $V = \{v_1, \dots, v_n\}$ derived from the CPC method can represent utilities, weights, scales, or priorities. A comparison score $b_{ij}$ measures the subjective difference between a pair of object values, approximated as $b_{ij} \cong v_i - v_j$. Completely measuring $n$ objects requires $\frac{n(n-1)}{2}$ pairwise comparisons. These scores form a subjective Pairwise Opposite Matrix $\mathbf{B} = [b_{ij}]$ based on paired interval scales, as defined below:

$$B = [b_{ij}] = \begin{bmatrix} 0 & b_{12} & \cdots & b_{1n} \\ b_{21} & 0 & \cdots & b_{2n} \\ \vdots & \vdots & \ddots & \vdots \\ b_{n1} & b_{n2} & \cdots & 0 \end{bmatrix} \quad (1)$$

The comparison score $b_{ij}$ reflects a subjective human judgment estimating the magnitude of difference between two paired objects. The values for $b_{ij}$ are selected from a paired interval scale in numerical form ($\Upsilon$) such that $b_{ij} \in \bar{\Upsilon} = \{\gamma_i\}$, where $\gamma_i$ is mathematically defined below:

$$\gamma_i = \frac{i\kappa}{8}, i = -1, \ldots, -|\gamma|, 0, |\gamma|, \ldots, 1 \quad (2)$$

Here, the parameter $\kappa$ (kappa) represents the root value for each measured object, and the magnitude level $i$ extends up to eight in either direction ($i = -8, \ldots, 8$). The recommended valuation schema is presented in Table 2. In this study, $\kappa$ is set to 8 if $2 \leq n \leq 8$, and $\kappa = n$ if $n > 8$.

Table 2. Paired Interval Scale schema for pairwise opposite comparison

| Level ($i$) | Numerical form ($\bar{\Upsilon}$) | Verbal form ($\gamma$) | Definition (Object A v.s. Object B) |
|---|---|---|---|
| 0 | 0 | Equally | Object A and object B are equal |
| 1 | $\frac{\kappa}{8}$ | Slightly | Object A is slightly more than object B |
| 2 | $\frac{2\kappa}{8}$ | Moderately | Object A is moderately more than object B |
| 3 | $\frac{3\kappa}{8}$ | Fairly | Object A is fairly more than object B |
| 4 | $\frac{4\kappa}{8}$ | Highly | Object A is highly more than object B |
| 5 | $\frac{5\kappa}{8}$ | Strongly | Object A is strongly more than object B |
| 6 | $\frac{6\kappa}{8}$ | Significantly | Object A is Significantly more than object B |
| 7 | $\frac{7\kappa}{8}$ | Outstandingly | Object A is outstandingly more than object B |
| 8 | $\kappa$ | Absolutely | Object A is absolutely more than object B |
| $L$ | Opposite values | | Object A is less than Object B in some degree |

Given the input matrix of comparison scores $[b_{ij}]$, the set of measured values $\{v_i\}$ is derived using the approximation $b_{ij} \cong v_i - v_j$. The matrix $[b_{ij}]$ consists of three structural components: an upper triangular matrix, a diagonal matrix, and a lower triangular matrix. The human evaluator provides comparison scores only for the upper triangular matrix, defined as $B^+ = \{b_{ij}: i < j, \forall i, \forall j\}$. The lower triangular matrix is the skew-symmetric inverse of the upper triangular matrix, represented as $B^- = \{b_{ij} = -b_{ji}: i > j, \forall i, \forall j\}$. When an object is compared against itself, the diagonal elements of $B$ are strictly zero, $b_{ij} = v_i - v_j = 0$ for $i = j$. The consistency of $B$ is validated using the Accordance Index (AI) below:

$$AI = \frac{1}{n^2} \sum_{i=1}^{n} \sum_{j=1}^{n} \sqrt{\frac{1}{n} \sum_{p=1}^{n} \left( \frac{(b_{ip} + b_{pj} - b_{ij})}{\kappa} \right)^2} \quad (3)$$

The $AI$ is a non-negative metric ($AI \geq 0$). If $AI = 0$, the matrix $B$ is perfectly accordant ($\tilde{B} \equiv B$). A value of $0 < AI \leq 0.1$ indicates a satisfactory level of consistency, whereas $AI > 0.1$ is considered unsatisfactory. Once $\{v_i\}$ is derived from $B$, the ideal pairwise opposite matrix $B$ is constructed as $\tilde{B} = [\tilde{b}_{ij}] = [v_i - v_j]$. This relationship is explicitly expressed as below:

$$\tilde{B} = \begin{bmatrix} 0 & v_1 - v_2 & \dots & v_1 - v_n \\ v_2 - v_1 & 0 & \dots & v_2 - v_n \\ \vdots & \vdots & \ddots & \vdots \\ v_n - v_1 & v_n - v_2 & \cdots & 0 \end{bmatrix} \cong \begin{bmatrix} 0 & b_{12} & \dots & b_{1n} \\ b_{21} & 0 & \dots & b_{2n} \\ \vdots & \vdots & \ddots & \vdots \\ b_{n1} & b_{n2} & \cdots & 0 \end{bmatrix} = B \tag{4}$$

The pairwise opposite matrix $B$ is reduced to a vector of object values $V$ via a cognitive reduction operator, denoted as $\psi: B \rightarrow V$. In other words, a matrix of paired interval scores is converted into a vector of individual values. The Primitive Least Squares (PLS) optimization model serves as a standard cognitive reduction operator below:

$$\begin{aligned} &\text{Min } \bar{\Delta} = \sum_{i=1}^{n} \sum_{j=i+1}^{n} \left(b_{ij} - v_i + v_j\right)^2 \\ &\text{s.t. } \sum_{i=1}^{n} v_i = n\kappa \end{aligned} \tag{5}$$

The objective function of the PLS model is to minimize the total squared difference between the upper triangular elements of $B$ and $B$, subject to the constraint that the summation of all object values equals $n\kappa$. The closed-form solution to this PLS optimization model is the Row Average plus the normal Utility (RAU) below:

$$RAU(B, \kappa) = \left[ v_i : v_i = \left( \frac{1}{n} \sum_{j=1}^{n} b_{ij} \right) + \kappa, \forall i \in \{1, \dots, n\} \right] \tag{6}$$

For most decision problems, the summation of weights $W = \{w_1, \dots, w_n\}$ is constrained to equal one ($\sum_{i=1}^{n} w_i = 1$), where $W$ represents a vector of normalized object values. The vector of individual object values $V$ is rescaled into this normalized weight vector using a standard normalization function below:

$$W = f_n(V) = \frac{V}{n\kappa} \tag{7}$$

## 5. CPC for HCAD

### 5.1. Fundamental definitions

A raw Weighted Data Table (WDT), as presented in Table 3, comprises a matrix of feature values $\{x_{ij}\}$ corresponding to a vector of observations $x = \{x_1, \dots, x_i, \dots, x_M\}$, a vector of target features $Y = \{Y_1, \dots, Y_j, \dots, Y_N\}$, and a vector of feature variables $X = \{X_1, \dots, X_j, \dots, X_N\}$. Here, $X_j = \{x_{\cdot,j}\}$ represents the vector of all observations for a specific feature variable $j$, while $x_i = \{x_{i,\cdot}\}$ represents the vector of all feature values for observation $i$. The individual value $x_{ij}$ may be either numerical or categorical.

Table 3. A weighted data table

| | $X_1$ | $\cdots$ | $X_j$ | $\cdots$ | $X_N$ | $Y$ |
|---|---|---|---|---|---|---|
| | $w_1$ | $\cdots$ | $w_j$ | $\cdots$ | $w_N$ | |
| $x_1$ | $x_{11}$ | $\cdots$ | $x_{1j}$ | $\cdots$ | $x_{1N}$ | $y_1$ |
| $\vdots$ | $\vdots$ | $\cdots$ | | $\cdots$ | $\vdots$ | $\vdots$ |
| $x_i$ | $x_{i1}$ | $\cdots$ | $x_{ij}$ | $\cdots$ | $x_{iN}$ | $y_i$ |
| $\vdots$ | $\vdots$ | $\cdots$ | | $\cdots$ | $\vdots$ | $\vdots$ |
| $x_M$ | $x_{M1}$ | $\cdots$ | $x_{Mj}$ | $\cdots$ | $x_{MN}$ | $y_M$ |

A raw dataset is rarely ready for direct modelling without prior data preprocessing, which typically encompasses discretization and categorical encoding. Subsequently, highly relevant features are selected for data modeling using machine learning algorithms. Feature weights can be incorporated alongside the data as inputs to these algorithms, directly influencing analytical outcomes and enhancing prediction or recommendation accuracy. Model selection encompasses both choosing an appropriate algorithm and configuring its parameter settings. The following subsections detail and exemplify CPC methods for weight allocation, categorical encoding, discretization (or categorization), holdout validation, feature selection, learning model selection, and weighted distances.

### 5.2. Feature Weighting with CPC

Features hold varying degrees of importance from the user's perspective. The Pairwise Opposite Matrix (POM) for feature set $X$, denoted by $B_X$, is constructed by evaluating each pair of feature variables. This process derives a vector of feature weights (or importance values) $\Lambda = \{\lambda_1, \dots, \lambda_j, \dots, \lambda_N\}$ that corresponds directly to the vector of feature variables $X = \{X_1, \dots, X_j, \dots, X_N\}$. The vector of importance values for the features is calculated as follows:

$$\Lambda = RAU(B_X, \kappa_X) \tag{8}$$

Compare the relative importance for each pair of the criteria below for movie recommendation. Circle one mark only for each pair.
X1: Movie Age; X2: Genres; X3: Number of Raters; X4: Average of Rating Scores; X5: Average of rating periods

| Criteria | Absolutely | Outstandingly | Significantly | Strongly | Highly | Fairly | Moderately | Slightly | Equally | Slightly | Moderately | Fairly | Highly | Strongly | Significantly | Outstandingly | Absolutely | Criteria |
|---|---|---|---|---|---|---|---|---|---|---|---|---|---|---|---|---|---|
| X1 | 8 | 7 | 6 | 5 | 4 | 3 | 2 | **1** | 0 | 1 | 2 | 3 | 4 | 5 | 6 | 7 | 8 | X2 |
| X1 | 8 | 7 | 6 | 5 | **4** | 3 | 2 | 1 | 0 | 1 | 2 | 3 | 4 | 5 | 6 | 7 | 8 | X3 |
| X1 | 8 | 7 | 6 | 5 | 4 | **3** | 2 | 1 | 0 | 1 | 2 | 3 | 4 | 5 | 6 | 7 | 8 | X4 |
| X1 | 8 | 7 | 6 | 5 | 4 | 3 | 2 | 1 | 0 | **1** | 2 | 3 | 4 | 5 | 6 | 7 | 8 | X5 |
| X2 | 8 | 7 | 6 | 5 | 4 | 3 | 2 | 1 | 0 | 1 | 2 | **3** | 4 | 5 | 6 | 7 | 8 | X3 |
| X2 | 8 | 7 | 6 | 5 | 4 | 3 | 2 | 1 | 0 | 1 | 2 | 3 | 4 | 5 | **6** | 7 | 8 | X4 |
| X2 | 8 | 7 | **6** | 5 | 4 | 3 | 2 | 1 | 0 | 1 | 2 | 3 | 4 | 5 | 6 | 7 | 8 | X5 |
| X3 | 8 | 7 | 6 | 5 | **4** | 3 | 2 | 1 | 0 | 1 | 2 | 3 | 4 | 5 | 6 | 7 | 8 | X4 |
| X3 | 8 | 7 | 6 | 5 | 4 | 3 | 2 | **1** | 0 | 1 | 2 | 3 | 4 | 5 | 6 | 7 | 8 | X5 |
| X4 | 8 | **7** | 6 | 5 | 4 | 3 | 2 | 1 | 0 | 1 | 2 | 3 | 4 | 5 | 6 | 7 | 8 | X5 |

Figure 3. The CPC survey form for comparisons of five features

Depending on the specific application, features may be organized into a hierarchical tree structure, an example of which is illustrated in Figure 4. Specifically, a parent feature $X_j$ within $X$ may encompass a set of sublevel features $\left(X_{j,1}, \dots, X_{j,k}, \dots, X_{j,N_k}\right)$. The Cognitive Pairwise Comparison (CPC) method can be recursively applied to evaluate these subtree features. Example 1 demonstrates how CPC is utilized to determine overall feature importance.

Table 4. Computational steps by RAU for priority vector of POM, $B_X$.

| $B_X$ | $X_1$ | $X_2$ | $X_3$ | $X_4$ | $X_5$ | | $\frac{1}{n}\sum_{j=1}^{n} b_{ij}$ | $v_i = \left(\frac{1}{n}\sum_{j=1}^{n} b_{ij}\right) + \kappa$ | $V' = \frac{V}{n\kappa}$ |
|---|---|---|---|---|---|---|---|---|---|
| $X_1$ | 0 | 1 | 4 | -1 | 6 | | 2 | 10 | 0.25 |
| $X_2$ | -1 | 0 | 3 | -3 | 4 | | 0.6 | 8.6 | 0.215 |
| $X_3$ | -4 | -3 | 0 | -6 | 1 | | -2.4 | 5.6 | 0.14 |
| $X_4$ | 1 | 3 | 6 | 0 | 7 | | 3.4 | 11.4 | 0.285 |
| $X_5$ | -6 | -4 | -1 | -7 | 0 | | -3.6 | 4.4 | 0.11 |
| | | AI=0.043 | | | | Sum: | 0 | 40 | 1 |

**Example 1**

An open-source dataset (MovieLens, 2019) is utilized to build a movie recommendation system. A user is asked to evaluate five features based on their definitions within the dataset. The CPC survey form utilized to compare these five features is illustrated in Figure 3. As shown in Table 4, the paired results from the CPC survey are structured into a POM, which is subsequently processed using the RAU operator (Eq. 8) to yield the final feature weights. The computational results indicate that the feature ratingScoreMean (RSM, $X_4$) emerges as the most important factor (weight = 0.285) from the user's perspective, followed by movieAge ($X_1$) (0.250), genres ($X_2$) (0.215), raterSize ($X_3$) (0.140), and ratingPeriodMean ($X_5$) (0.110).

### 5.3. Categorical Encoding with CPC

Categorical encoding is defined as a mapping function $f_{enc}: H^{(j)} \rightarrow G^{(j)}$ that assigns a vector of numerical values $G^{(j)} = \left\{G_1^{(j)}, G_2^{(j)}, \ldots, G_t^{(j)}, \ldots, G_T^{(j)}\right\}$ to a corresponding vector of categorical labels $H^{(j)} = \left\{H_1^{(j)}, H_2^{(j)}, \ldots, H_t^{(j)}, \ldots, H_T^{(j)}\right\}$ for a feature variable $j$. The Cognitive Pairwise Comparison (CPC) method for categorical encoding is formulated as follows:

$$G^{(j)} = RAU\left(B_{enc}^{(j)}, \kappa_{enc}^{(j)}\right) \cdot s \tag{9}$$

where $B_{enc}^{(j)}$ represents a Pairwise Opposite Matrix (POM) evaluating user preferences for categorical labels in feature $j$ (denoted by $H^{(j)}$), and $\kappa_{enc}^{(j)}$ is the root utility parameter (typically the mean) for presenting $H^{(j)}$. The scaling factor $s$ is set to 1 by default, or $s = T$ if the number of compared objects $T$ is large. The output vector derived from the RAU function taking $B_{enc}^{(j)}$ and tuning parameter $\kappa_{enc}^{(j)}$ is directly applied to encode the categorical labels. When dummy variables are created via one-hot encoding of multiple labels, an observation record associated with a label attribute in $X_{j,k}$ (denoted as $x_{i,(j,k)}$) takes the corresponding preference weight instead of standard binary values:

$$x_{i,(j,k)} \in \left\{G_1^{(j)}, \ldots, G_{N_k}^{(j)}\right\} \tag{10}$$

Equivalently, weights are allocated directly across the set of dummy variables:

$$X_j = \{X_{j,1}, \ldots, X_{j,k}, \ldots, X_{j,N_k}\} = \left\{G_1^{(j)}, \ldots, G_{N_k}^{(j)}\right\} \tag{11}$$

These weighted dummy variables are subsequently multiplied by the binary matrix of dummy variables, where a value of 1 indicates that an observation record is associated with a given label attribute. Algorithm 1 outlines the categorical encoding steps, and Example 2 demonstrates how CPC is utilized to perform categorical encoding in practice.

**Example 2**

Eighteen attributes for movie genres $X_2$ (shown in Table 9 and Figure 4) are transformed into dummy variables $X_{2,1}, \ldots, X_{2,18}$. While a user evaluates their preferences using CPC, the POM for CPC categorical encoding ($B_{enc}^{(2)}$) is constructed as presented in Table 5. The Accordance Index (AI) of $B_{enc}^{(2)}$ is 0.076; thus, $B_{enc}^{(2)}$ satisfies consistency requirements. The last row of Table 5 displays the user's perceived preference weights ($G^{(2)}$) across the eighteen genre attributes.

For example, as shown in Table A1 in the Appendix, the movie *Jumanji (1995)* (movieId = 2) includes Adventure ($X_{2,1}$), Children ($X_{2,3}$), and Fantasy ($X_{2,5}$), which are assigned weights of 1.55, 0.85, and 1.58, respectively, instead of binary ones as in conventional one-hot encoding, while all remaining dummy variables created from genres are assigned zero.

Table 5. POM for categorical encoding of genre attributes

| | $X_{2,1}$ | $X_{2,2}$ | $X_{2,3}$ | $X_{2,4}$ | $X_{2,5}$ | $X_{2,6}$ | $X_{2,7}$ | $X_{2,8}$ | $X_{2,9}$ | $X_{2,10}$ | $X_{2,11}$ | $X_{2,12}$ | $X_{2,13}$ | $X_{2,14}$ | $X_{2,15}$ | $X_{2,16}$ | $X_{2,17}$ | $X_{2,18}$ |
|---|---|---|---|---|---|---|---|---|---|---|---|---|---|---|---|---|---|---|
| $X_{2,1}$ | 0 | 3 | 6 | 8 | 0 | 7 | 7 | 1 | 8 | 3 | 4 | 0 | 1 | 5 | 2 | 8 | 8 | 8 |
| $X_{2,2}$ | -3 | 0 | 3 | 6 | -3 | 5 | 4 | -1 | 7 | 1 | 2 | -2 | -2 | 3 | -1 | 6 | 6 | 7 |
| $X_{2,3}$ | -6 | -3 | 0 | 3 | -6 | 2 | 1 | -5 | 4 | -3 | -2 | -6 | -5 | -1 | -4 | 2 | 3 | 4 |
| $X_{2,4}$ | -8 | -6 | -3 | 0 | -8 | -1 | -2 | -7 | 1 | -5 | -4 | -8 | -8 | -3 | -7 | 0 | 0 | 1 |
| $X_{2,5}$ | 0 | 3 | 6 | 8 | 0 | 8 | 7 | 2 | 8 | 4 | 5 | 1 | 1 | 5 | 2 | 8 | 8 | 8 |
| $X_{2,6}$ | -7 | -5 | -2 | 1 | -8 | 0 | -1 | -6 | 2 | -4 | -3 | -7 | -7 | -2 | -6 | 1 | 2 | 2 |
| $X_{2,7}$ | -7 | -4 | -1 | 2 | -7 | 1 | 0 | -5 | 3 | -3 | -2 | -6 | -6 | -2 | -5 | 1 | 2 | 3 |
| $X_{2,8}$ | -1 | 1 | 5 | 7 | -2 | 6 | 5 | 0 | 8 | 2 | 3 | -1 | 0 | 4 | 1 | 7 | 8 | 8 |
| $X_{2,9}$ | -8 | -7 | -4 | -1 | -8 | -2 | -3 | -8 | 0 | -6 | -5 | -8 | -8 | -4 | -8 | -1 | 0 | 0 |
| $X_{2,10}$ | -3 | -1 | 3 | 5 | -4 | 4 | 3 | -2 | 6 | 0 | 1 | -3 | -2 | 2 | -1 | 5 | 6 | 6 |
| $X_{2,11}$ | -4 | -2 | 2 | 4 | -5 | 3 | 2 | -3 | 5 | -1 | 0 | -4 | -3 | 1 | -2 | 4 | 5 | 5 |
| $X_{2,12}$ | 0 | 2 | 6 | 8 | -1 | 7 | 6 | 1 | 8 | 3 | 4 | 0 | 0 | 5 | 2 | 8 | 8 | 8 |
| $X_{2,13}$ | -1 | 2 | 5 | 8 | -1 | 7 | 6 | 0 | 8 | 2 | 3 | 0 | 0 | 4 | 1 | 7 | 8 | 8 |
| $X_{2,14}$ | -5 | -3 | 1 | 3 | -5 | 2 | 2 | -4 | 4 | -2 | -1 | -5 | -4 | 0 | -3 | 3 | 4 | 5 |
| $X_{2,15}$ | -2 | 1 | 4 | 7 | -2 | 6 | 5 | -1 | 8 | 1 | 2 | -2 | -1 | 3 | 0 | 6 | 7 | 8 |
| $X_{2,16}$ | -8 | -6 | -2 | 0 | -8 | -1 | -1 | -7 | 1 | -5 | -4 | -8 | -7 | -3 | -6 | 0 | 1 | 2 |
| $X_{2,17}$ | -8 | -6 | -3 | 0 | -8 | -2 | -2 | -8 | 0 | -6 | -5 | -8 | -8 | -4 | -7 | -1 | 0 | 1 |
| $X_{2,18}$ | -8 | -7 | -4 | -1 | -8 | -2 | -3 | -8 | 0 | -6 | -5 | -8 | -8 | -5 | -8 | -2 | -1 | 0 |
| $G^{(2)}$ | 1.55 | 1.26 | 0.85 | 0.53 | 1.58 | 0.65 | 0.75 | 1.42 | 0.44 | 1.17 | 1.05 | 1.52 | 1.47 | 0.94 | 1.35 | 0.57 | 0.48 | 0.42 |

### 5.4 Discretisation (Binning or Categorisation) with CPC

The scale of measurement includes nominal, ordinal, interval, and ratio scales. Nominal and ordinal scale data represent categorical data, whereas interval and ratio data represent numerical data. Nominal data possess no intrinsic ordering, whereas ordinal data follow a defined sequence. Interval data feature an arbitrary zero point, whilst ratio data possess a natural zero point. Some machine learning methods—such as conventional association rule mining—require purely categorical data as inputs. Discretization (also termed binning or categorization) is the process of transforming continuous numerical data into discrete categorical variables. Cognitive Pairwise Comparisons are extended to numerical discretization to transform data formats to satisfy the input requirements of specialized machine learning algorithms.

Let $G^{(j)}$ denote the domain of numerical values for feature variable $j$, and let $H^{(j)}$ denote the domain of discrete categorical values for feature variable $j$. Discretization is defined as a mapping function $f_{dsc}: G^{(j)} \to H^{(j)}$ that transforms numerical values into their categorical counterparts for feature variable $j$. A set of $T-1$ cut points $\{h_t^{(j)}: t = 1, \ldots, T-1\}$ for feature $j$ establishes $T$ interval boundaries for the categorical labels, defined as follows:

$$\begin{aligned} H^{(j)} &= \left\{H_1^{(j)}, H_2^{(j)}, \ldots, H_t^{(j)}, \ldots, H_T^{(j)}\right\} \\ &= \left\{\left(-\infty, h_1^{(j)}\right), \left[h_1^{(j)}, h_2^{(j)}\right), \ldots, \left[h_{t-1}^{(j)}, h_t^{(j)}\right), \ldots, \left[h_{T-1}^{(j)}, \infty\right)\right\} \end{aligned} \quad (12)$$

Here, $H_t^{(j)} = \left[h_{t-1}^{(j)}, h_t^{(j)}\right)$ for $t = 2, \ldots, T$ represents categorical label $t$ bounded between exclusive lower cut point $h_{t-1}^{(j)}$ and inclusive upper cut point $h_t^{(j)}$ for transformed feature $j$. Any value strictly less than cut point $h_1^{(j)}$ is assigned to $H_1^{(j)}$, whereas any value greater than or equal to cut point $h_{T-1}^{(j)}$ is assigned to $H_T^{(j)}$ (i.e., $h_0^{(j)} = -\infty$ and $h_T^{(j)} = \infty$).

The CPC method is adopted to specify interval widths that define cut points for converting numerical values into categorical counterparts. For a numerical feature $j$, the proportions across $T$ interval boundaries are calculated using CPC via the Normalized Row Average plus normal Utility (NRAU) operator as follows:

$$W = \{w_1, \dots, w_T\} = NRAU\left(B_{dsc}^{(j)}, \kappa_{dsc}^{(j)}\right) \quad (13)$$

The numerical domain is partitioned into $T$ intervals according to the width proportions vector $W$. The cut points for the interval boundaries of feature $j$ (denoted by $h_t^{(j)}$ for $\forall t \in \{1, \dots, T-1\}$) are computed using as below:

$$h_t^{(j)} = \delta \cdot min(X_j) + \left\{\left(max(X_j) - min(X_j)\right)\sum_{i=1}^{t} w_i\right\}, \forall t \in \{1, \dots, T-1\} \quad (14)$$

Numerical interval values increment sequentially from the minimum value $\min(X_j)$. The total range of the interval is $\left(\max(X_j) - \min(X_j)\right)$. The weight vector $W$ dynamically adjusts interval widths evenly or unevenly according to user preferences. The parameter $\delta$ is a scaling factor that adjusts the starting point to ensure that $h_{T-1}^{(j)} < \max(X_j)$ (by default, $\delta = 1$). Because this formulation is sensitive to extreme minimum values, $\delta$ can be adjusted to mitigate the impact of outlier starting points.

**Example 3**

User rating scores serve as target features for classification models to evaluate and predict movie quality. However, target values across movies (ranging from 0.5 to 5.0) represent continuous average numerical ratings aggregated across users. The average rating scores for all movies (X3) are categorized into three ordinal labels: *poor*, *fair*, and *good*, as structured in Table 9 and Figure 5. Table 6 details the computational steps of CPC discretization for the rating score target feature. Additionally, Table 16 illustrates the discretization procedure used to partition movie release years into four distinct categories.

Table 6: Discretisation with CPC for movie rating scores

| $\delta = 1$, $\min(X_3) = 0.5$, $\max(X_3) = 5$, AI = 0.048 | | | | | | | |
|---|---|---|---|---|---|---|---|
| index | $B_{dsc}^{(3)}$ | poor | fair | good | $w_i$ | $h_t^{(3)}$ | $H_t^{(3)}$ |
| 1 | poor | 0 | 3 | 4 | 0.431 | 2.438 | [-∞,2.438) |
| 2 | fair | -3 | 0 | 2 | 0.319 | 3.875 | [2.438, 3.875) |
| 3 | good | -4 | -2 | 0 | 0.250 | | [3.875,∞] |

## 5.5 Training-Validation-Test Data Split with CPC

For supervised learning, a dataset is typically partitioned into two or three distinct subsets: a training set ($X_{trn}$), a validation set ($X_{val}$), and a testing set ($X_{tst}$), where the validation subset may be omitted when simpler sampling strategies suffice. Cognitive Pairwise Comparisons (CPC) can be systematically employed to specify proportional ratios for partitioning the dataset. Algorithm 1 details the procedural steps to split a dataset into training, validation, and testing subsets using the CPC methodology. An empirical demonstration of this CPC data splitting procedure is presented in Table 10.

**Algorithm 1: Training-validation-test data split with CPC**

**Input / Function Signature:** $f_{tvt}(X, B_{tvt}, \kappa_{tvt})$

1. Compute numerical ratio values for the subset categories:
   $\{w_{trn}, w_{val}, w_{tst}\} = NRAU(B_{tvt}, \kappa_{tvt})$

2. Randomize row positions in dataset $X$:
   $\tilde{X} = Rnd(X, seed)$ *# set the seed number to ensure the same random pattern*

3. Determine index cut points for partitioning:
   $L_t = \lfloor M \sum_{i=1}^{t} w_i \rfloor, t = 1,2,3$

4. Allocate observations across subsets:
   $\tilde{X}_{trn} = \tilde{X}_{i=(1,\dots,L_1)}$
   $\tilde{X}_{val} = \tilde{X}_{i=(L_1+1,\dots,L_2)}$
   $\tilde{X}_{trn} = \tilde{X}_{i=(L_2+1,\dots,M)}$

**Return:** $\{\tilde{X}_{trn}, \tilde{X}_{val}, \tilde{X}_{tst}\}$

### 5.6 Features selection with CPC

Not all features in a dataset effectively contribute to data analytics modeling. Irrelevant features included during model training can introduce extra computational workload, unnecessary noise, and degraded predictive accuracy. Consequently, minimizing feature dimensionality while retaining critical signals represents a fundamental objective in building robust machine learning pipelines. In the proposed framework, feature relevance is established directly from user importance values elicited and quantified through Cognitive Pairwise Comparisons (CPC).

Feature selection is formulated as a function that selects an optimal subset from the full set of raw features to maximize analytical performance. The vector of feature weights $\Lambda = \{\lambda_1, \dots, \lambda_j, \dots, \lambda_N\}$ corresponding to the feature variables $X = \{X_1, \dots, X_j, \dots, X_N\}$ is derived using the Row Average plus the normal Utility (RAU) operator as follows:

$$\Lambda = RAU(B_X, \kappa_X) \tag{15}$$

Let $\bar{\Lambda} = \{\bar{\lambda}_1, \dots, \bar{\lambda}_i, \dots, \bar{\lambda}_N\}$ represent the vector of feature importance values sorted in strictly descending order from the original unsorted vector $\{\lambda_1, \dots, \lambda_j, \dots, \lambda_N\}$ such that $\bar{\lambda}_1 \geq \cdots \geq \bar{\lambda}_i \geq \cdots \geq \bar{\lambda}_N$. The general sorting operation for feature importance values and their corresponding variable indices is defined as follows:

$$\{(\vec{\lambda}_i, \theta_i)\} = f_{dsc}(\{\lambda_j, j\}) = f_{dsc}(\{\lambda_j\}, \{j\}) \tag{16}$$

Here, the importance values $\{\lambda_1, \dots, \lambda_j, \dots, \lambda_N\}$ are sorted alongside their original feature indices $\{j\}$, returning paired tuples $\{(\bar{\lambda}_i, \theta_i)\}$ ordered by user preferences. Once paired tuples ($\{\vec{\lambda}_i\}$,$\{\theta_i\}$) are determined, an optimal subset of best $n$ features is selected from the total $N$ candidate features.

Algorithm 2 outlines Forward Feature Selection with CPC (denoted as $f_{fsc}(X, B_X, \kappa_X, \psi)$). The algorithm computes the feature importance vector $\{\lambda_j\}$ using CPC and derives the ordered variable index sequence $\{\theta_i\}$. Following this prioritized sequence, features are candidate-tested and iteratively appended to the active feature subset whenever they yield performance improvements. The evaluation function $\psi(\cdot)$ represents a dedicated domain metric measuring predictive or explanatory performance, defined as below.

$$\psi(\beta^+) = \begin{cases} \psi_c, if\ c =' +' \\ -\psi_c, if\ c =' -' \end{cases} \tag{17}$$

Evaluation criteria vary across machine learning paradigms. Negative metrics denoted by '−', (e.g., Mean Squared Error or Root Mean Squared Error) indicate superior performance at lower values, whereas positive metrics (denoted by '+' ), e.g., Accuracy, Precision, or F1-Score) indicate superior performance at higher values. The subset $\beta^+$ represents the active dataset containing the selected and ordered features. Upon completing Algorithm 2, the optimal feature subset and index mapping $\{\beta^+, \theta^+\}$ are returned.

**Algorithm 2: Forward features selection with training-validation-test split and CPC**

**Input / Function Signature:** $f_{fsc}(X, B_X, \kappa_X, \psi, B_{tvt}, \kappa_{tvt})$

1. Split the complete dataset into Training, Validation, and Testing Subsets:

   $\{\tilde{X}_{trn}, \tilde{X}_{val}, \tilde{X}_{tst}\} = f_{tvt}(X, B_{tvt}, \kappa_{tvt})$
2. Calculate feature importance weights across hierarchical levels:

   $\{\lambda_j\} = RAU(B_X, \kappa_X)$
3. Extract ordered variable indices by sorting importance weights $\{\lambda_j\}$ in descending order

   $\{\theta_i\} = f_{dsc}(\{\lambda_j\}, \{j\})$
4. Initialize the parameters:

   $\beta^+_{trn} = \{1\}$

   $\beta^+_{val} = \{1\}$

   $\theta^+ = \{\theta_0\}$
5. Add additional features one by one when improvement is added

   For $i = 1$ to $N$ do

   $\beta^+_{trn} = \beta^+_{trn} \cup \tilde{X}^{\theta_i}_{trn}$ # Add one additional feature variable

   $\beta^+_{trn} = \beta^+_{trn} \cup \tilde{X}^{\theta_i}_{trn}$

   If $\psi(\beta^+_{trn}) > \psi(\beta_{trn})$ & $\psi(\beta^+_{val}) > \psi(\beta_{val})$ # If training and validation performance improve

   $\beta^+ = \beta$ # Accept and update feature subset

   $\theta^+ = \theta^+ \cup \theta_i$

**Return:** $\{\beta^+, \theta^+\}$

An empirical application of CPC forward feature selection is demonstrated in Table 11 (Section 6.3). Algorithm 2 is specifically designed for standard machine learning pipelines, seamlessly integrating dataset partitioning (Algorithm 1) with preference-guided forward feature selection. Because variable ordering heavily dictates feature selection trajectories and overall model generalization, the CPC methodology provides a mathematically sound, soft-computing foundation to encode human domain expertise directly into algorithmic decision-making.

### 5.7 Learning models selection with CPC decision-making model

Although more than one algorithm can be used for machine learning or data analytics tasks, selecting the best algorithm for a task can be challenging, as everyone claims that their methods are the best. When evaluating the best algorithm, more than one benchmark function can be used to measure algorithm candidates. To create a structured decision problem, let $\{Y_1, \dots, Y_i, \dots, Y_P\}$ be $P$ learning methods evaluated by the $Q$ benchmarking methods $\{\psi_1, \dots, \psi_j, \dots, \psi_Q\}$ with the criteria weights $\{\pi_1, \dots, \pi_j, \dots, \pi_Q\}$.

Table 7: CPC weighted decision matrix for learning models selection

| | $\psi_1$ | $\cdots$ | $\psi_j$ | $\cdots$ | $\psi_Q$ | Z |
|---|---|---|---|---|---|---|
| | $\pi_1$ | $\cdots$ | $\pi_j$ | $\cdots$ | $\pi_Q$ | |
| $R_1$ | $r_{11}$ | $\cdots$ | $r_{1j}$ | $\cdots$ | $r_{1Q}$ | $z_1$ |
| $\vdots$ | $\vdots$ | $\cdots$ | $\vdots$ | $\cdots$ | $\vdots$ | $\vdots$ |
| $R_i$ | $r_{i1}$ | $\cdots$ | $r_{ij}$ | $\cdots$ | $r_{iQ}$ | $z_i$ |
| $\vdots$ | $\vdots$ | $\cdots$ | $\vdots$ | $\cdots$ | $\vdots$ | $\vdots$ |
| $R_P$ | $r_{P1}$ | $\cdots$ | $r_{Pj}$ | $\cdots$ | $r_{PQ}$ | $z_P$ |

The POM for $\psi$, denoted by $B_\psi$, is formed by evaluating all pairs of benchmarking methods. After $B_\psi$ is reduced by the function NRAU, the criteria weights of the benchmarking methods are calculated as follows:

$$\{\pi_i\} = NRAU(B_\psi, \kappa_\psi) \tag{18}$$

The CPC-weighted decision matrix for the learning model selection is shown in Table 7. Once the benchmarking values $\{r_{ij}\}$ for all learning models with respect to all benchmarking methods are obtained, $\{r_{ij}\}$ and $\{\pi_j\}$ are aggregated by the aggregation function below:

$$Z = f_{agg}(\{\pi_j\}, \{r_{ij}\}) \tag{19}$$

The weighted mean below is the aggregation function used in this study, as it is well-known, efficient, and widely acceptable:

$$z_i = \sum_{j=1}^{Q} \pi_j r_{ij}\,, \qquad i = 1, \dots, P \tag{20}$$

The learning model with the best $z$ value was chosen among all the candidate learning models. An example is presented in Tables 13 and 14 in Section 6.4.

### 5.8 Weighted distances with CPC

Similarity or dissimilarity between two objects or observations is measured by their feature-weighted distance. Euclidean distance is a standard metric for distance measurement that assumes equal weights across all features. When feature weights are unequal, the weighted Euclidean distance between observation $i$ and observation $k$ is defined as follows:

$$wed_{ik} = \sqrt{\sum_j w_j \left(x_{ij} - x_{kj}\right)^2}, w_j \in W \tag{21}$$

Cognitive Pairwise Comparisons (CPC) is utilized to elicit user preferences and derive the normalized feature weight vector $W$, using the Row Average plus the normal Utility (RAU) operator:

$$W = RAU(B_X, \kappa_X) \quad (22)$$

The following theorem establishes an essential computational property of the weighted Euclidean distance, facilitating its integration into existing analytics algorithms.

**Theorem 1**: For a pair of objects represented in rows i and k across N features in a data matrix X, let a weighted transformed data matrix $\dot{X}$ be defined by the element-wise multiplication of the row vector of square-root weights $\sqrt{W}$ and data matrix X—that is, $\dot{X} = \sqrt{W}X = \{\dot{x}_{ij}\} = \{\sqrt{w_j}x_{ij}\}$. The Weighted Euclidean Distance between X and W is mathematically equivalent to the unweighted Euclidean Distance computed on $\dot{X}$—that is:

$$\begin{aligned} wed(W,X) &= \sqrt{\sum_j w_j\left(x_{ij} - x_{kj}\right)^2} \\ &= ed(\dot{X}) = \sqrt{\sum_j \left(\dot{x}_{ij} - \dot{x}_{kj}\right)^2}, i \neq k, \dot{x}_{ij}, \dot{x}_{kj} \in \dot{X} = \sqrt{W}\mathrm{X} \end{aligned} \quad (23)$$

Proof:

$$\sum_j w_j\left(x_{ij} - x_{kj}\right)^2 = \sum_j\left(\sqrt{w_j}\right)^2\left(x_{ij} - x_{kj}\right)^2 = \sum_j\left(\sqrt{w_j}x_{ij} - \sqrt{w_j}x_{kj}\right)^2. \quad \text{Q.E.D}$$

From a practical implementation standpoint, Theorem 1 demonstrates that incorporating user preferences into distance-based algorithms does not require modifying core algorithm source code. One simply pre-multiplies the data matrix $X$ by $\sqrt{W}$ to construct the weighted matrix $\dot{X}$, which can then be passed directly into standard software packages configured for unweighted Euclidean distance ($ed$). Thus:

$$wed(X, W) = ed(\dot{X}) \quad (24)$$

An empirical demonstration applying weighted Euclidean distances to evaluate the Weighted Average Silhouette Index (WASI) within a movie recommender clustering pipeline is detailed in Section 6.5.

### 5.9 User Preference Weighted Data Reduction with CPC

Table 8. Definitions of terms in principle components

| | |
|---|---|
| Loadings $\Gamma = V\sqrt{L}$ (or $\Gamma = L$ ) | Proportions of variances $L\grave{} = \frac{L}{\sum L}$ |
| Standard deviations $\Lambda = \sqrt{L}$ | Component Scores $\Theta = V\hat{Z}$ |

Data reduction is used to transform feature variables into principal components, for which the top components are retained for downstream modeling. Principal Component Analysis (PCA) is a widely established data reduction technique. When importance weights are assigned to feature variables, features with higher weights contribute more significantly to the variance captured by the top components. Cognitive Pairwise Comparisons (CPC) are employed to determine these user preference weights. Algorithm 3 presents user-preference-weighted data reduction using PCA with CPC. In this workflow, feature variables are standardized to have zero mean, thereby unifying the scale across raw features. The standardized feature values are multiplied by their respective preference weights to construct a weighted correlation matrix, which is subsequently decomposed into eigenvalues ($L$) and eigenvectors ($V$) via eigensystem decomposition. These eigenvalues $L$ and eigenvectors $V$ are further utilized to define component loadings, standard deviations, variance proportions, and component scores, as defined in Table 8.

Because top principal components are inherently sensitive to feature variances, variables exhibiting larger variances naturally dominate those with smaller variances in standard PCA. By assigning a higher preference weight to a feature, user-preference-weighted PCA amplifies that feature's variance, thereby increasing its relative contribution to the top principal components. These weighted components are subsequently utilized to build clustering models that uncover preference-aligned patterns. An empirical application of weighted data reduction is illustrated in Figure 10 (Section 6.5).

**Algorithm 3: User-preference-weighted PCA with CPC**

**Input / Function Signature:** $RPWDR(X, \{(B, \kappa)\})$

1. Standardize data features with zero mean and unit variance
   where $\mu_j$ and $\sigma_j$ represent the mean and standard deviation of feature $j$
   $$Z_j = \frac{X_j - \mu_j}{\sigma_j}, \forall j \in \{1, \dots, N\}$$

2. Calculate feature importance weights across hierarchical levels:
   $$W = \{w_j\} = HRAU(\{(B, \kappa)\}) \times N$$

3. Calculate element product of importance values and the standardised data
   $$\hat{Z} = W * Z$$

4. Calculate the weighted correlation matrix $\hat{R}$ for $\hat{Z}$ where $\hat{m}_i$ represents the mean of weighted standardized feature $i$ ( $\hat{Z}_i$).
   $$\hat{R}_{ij} = \frac{\sum(\hat{Z}_i - \hat{m}_i)(\hat{Z}_j - \hat{m}_j)}{\sqrt{\sum(\hat{Z}_i - \hat{m}_i)^2 \sum(\hat{Z}_j - \hat{m}_j)^2}}, \forall i, j \in \{1, \dots, N\}$$

5. Derive eigenvalues $L$ and eigenvectors $V$ for $\hat{R} = \{\hat{R}_{ij}\}$
   $$LV = \hat{R}V\ ,\ L_1 < \cdots < L_N$$

6. Compute output metrics defined in Table 7.

**Return**: { L,V, Θ, Γ , L`, Λ}

## 6. Human-Centric Movies Evaluation and Recommendation with CPC

### 6.1 Background and Motivations with CPC

To illustrate the practical applicability and robustness of the proposed Cognitive Pairwise Comparisons for Human-Centric Artificial Intelligence and Data Science (CPC-HCAD) framework, the development of human-centric movie recommendation applications is discussed in detail. The benchmark open-source MovieLens 25M dataset (MovieLens, 2019), constructed as described by Harper and Konstan (2015), is utilized to build machine learning models guided by CPC. This approach facilitates continuous interaction between human decision-makers and algorithmic pipelines within human-in-the-loop (HITL) processes. All processed datasets, computational web modules for core algorithms, and supplementary materials are accessible via the author's GitHub repository at https://github.com/kkfyuen/cpcHCADDemos and permanently archived on Zenodo at https://doi.org/10.5281/zenodo.23434993. Live computational web module demos are hosted on GitHub Pages at https://kkfyuen.github.io/cpcHCADDemos/. The key contributions of CPC across core AI&DS techniques are summarized as follows:

- Data transformation: Raw metadata cannot be fed directly into machine learning algorithms without prior preprocessing to meet strict algorithmic input parameters. Within this framework, CPC is systematically applied to perform feature weighting, categorical encoding, and continuous data discretization.
- Regression analysis: Variable preference ordering and representative data sampling represent critical prerequisites for effective regression modeling. Human-guided forward feature selection (Algorithm 2) and training-validation-test data partitioning (Algorithm 1) with CPC are deployed to address these requirements, systematically exploring feature spaces to predict continuous movie rating scores.
- Classification model selection: Selecting the optimal classification algorithm among multiple candidate models is challenging, as competing algorithms yield disparate performance profiles across multiple evaluation metrics. Model selection with CPC (Eqs. 18–20) is employed to evaluate, rank, and select the best predictive model to classify movie rating categories established via CPC discretization.

- Weighted data reduction and clustering: Different user groups prioritize movie features differently. CPC encodes subjective user preferences into weighted Euclidean distance metrics (Eqs. 21–24) and user-preference-weighted Principal Component Analysis (Algorithm 3), guiding K-Means clustering to discover preference-aligned viewer segments for personalized movie recommendations.
- Association rule learning: Continuous numerical features cannot be natively processed by traditional association rule mining algorithms. CPC discretization transforms continuous movie attributes into discrete categorical bins, which are subsequently converted into binary dummy variables. Frequent itemsets and association rules generated via the Apriori algorithm are then prioritized and ranked according to CPC preference weights to deliver tailored recommendation lists.

### 6.2 **Data transformation with CPC**

The empirical evaluation of the Cognitive Pairwise Comparisons for Human-Centric Artificial Intelligence and Data Science (CPC-HCAD) framework utilizes two core relational files from the open-source MovieLens 25M benchmark collection (MovieLens, 2019): *movies.csv* (containing 62,423 raw film entries across movieId, title, and genres) and *ratings.csv* (comprising 25,000,095 user evaluations submitted by 162,541 unique evaluators across 59,047 movies). Representative data snapshots are provided in Table A1 and Table A2 in the Appendix. The technical procedures and computational modules for this section are hosted via the following open-source web links:

- https://kkfyuen.github.io/cpcHCADDemos/s1dataTransformation.html
- https://kkfyuen.github.io/cpcHCADDemos/s2cpcTransformation.html

To prepare raw metadata for algorithmic modelling, a rigorous multi-stage cleaning and feature engineering pipeline was executed. First, release year patterns ($YYYY$) were systematically extracted from movie titles using regular expressions. A total of 406 entries missing valid release years (a missing rate of 0.65%) were excluded, leaving 62,017 valid movies. Historical release year anomalies were individually verified and corrected—for example, rectifying *1013 Briar Lane* (year 1013 → 2009) while validating legitimate early cinema records such as *Passage de Vénus* (1874). Raw title strings were subsequently cleaned by stripping parenthetical release years. Next, genre strings were parsed across 20 unique categories. Uninformative or invalid entries—specifically (no genres listed) (4,801 titles) and IMAX (1 title)—were removed (totalling 4,802 excluded titles), preserving 57,215 valid movies across 18 target genre categories.

Simultaneously, *ratings.csv* was aligned with the cleaned movie table, filtering out 36,723 ratings associated with excluded movies and retaining 24,963,372 ratings across 54,351 shared movies. Evaluator timestamps were converted to calendar years, and user rating evaluations were aggregated per movie to compute summary statistics: rater volume (*raterSize*), mean rating score (*ratingScoreMean*), standard deviation (*ratingScoreSD*), and average rating year (*ratingYearMean*). Two key temporal metrics were engineered: movie age ($movieAge = 2019 - year$) and average rating period duration ($ratingPeriodMean = ratingYearMean - year$). Records exhibiting negative rating duration periods (85 movies where user ratings were recorded prior to official film release years) were removed, yielding a final, fully sanitized dataset (*moviesDB*) containing exactly 54,266 unique movie records.

The sanitized dataset is structured into two distinct feature configurations tailored to specific machine learning paradigms, as illustrated in Figure 4, Figure 5, and Table 9.

- 22-Feature Continuous/Scaled Architecture (Regression, Classification, Clustering): Comprises 4 continuous/scaled predictor variables—*movieAge* ($X_1$), *raterSize* ($X_3$), *ratingPeriodMean* ($X_4$), and *ratingScoreMean* ($X_5$)—alongside 18 binary genre dummy variables ($X_{2,1},\ldots,X_{2,18}$) weighted via CPC categorical encoding ($B_{enc}^{(2)}$). For regression tasks, *ratingScoreMean* serves as the continuous target; for classification, it is discretized into 3 ordinal rating classes (poor, fair, good).
- 33-Feature Discretized Dummy Architecture (Association Rule Mining): To satisfy input constraints of transaction-based itemset mining algorithms, continuous attributes are discretized using CPC preference-proportional binning (Eqs. 12–14) into 33 binary dummy variables (4 + 18 + 4 + 3 + 4 = 33): year (4 categories: Early, Classical, Modern, Recent), genres (18 categories), raterSize (4 categories: little, some, enough, plenty), Grade (3 categories: poor, fair, good), and ratingPeriodMean (4 categories: recent, quite recent, long time ago, very long time ago).

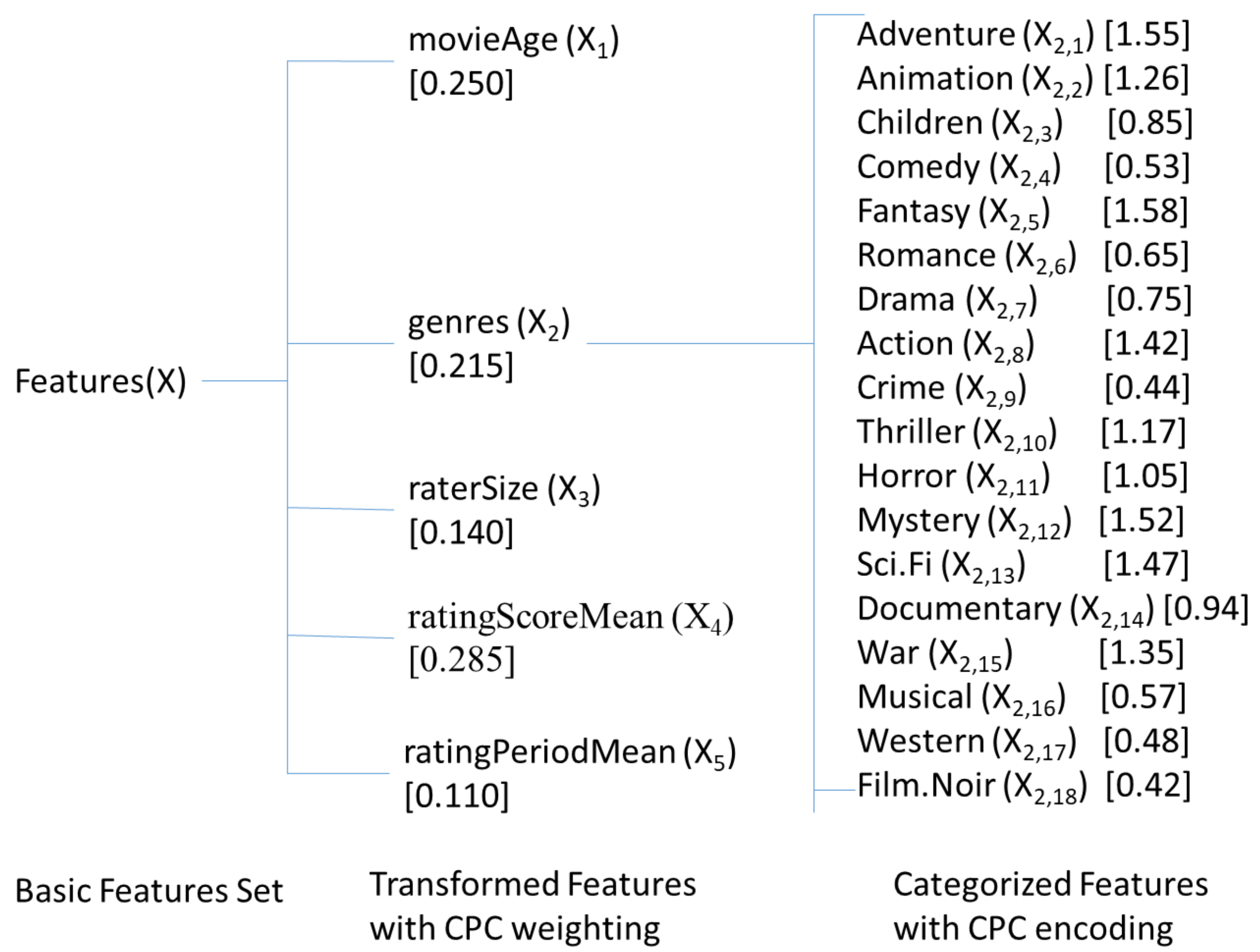

Figure 4: Transformed features with CPC encoding Categorical feature

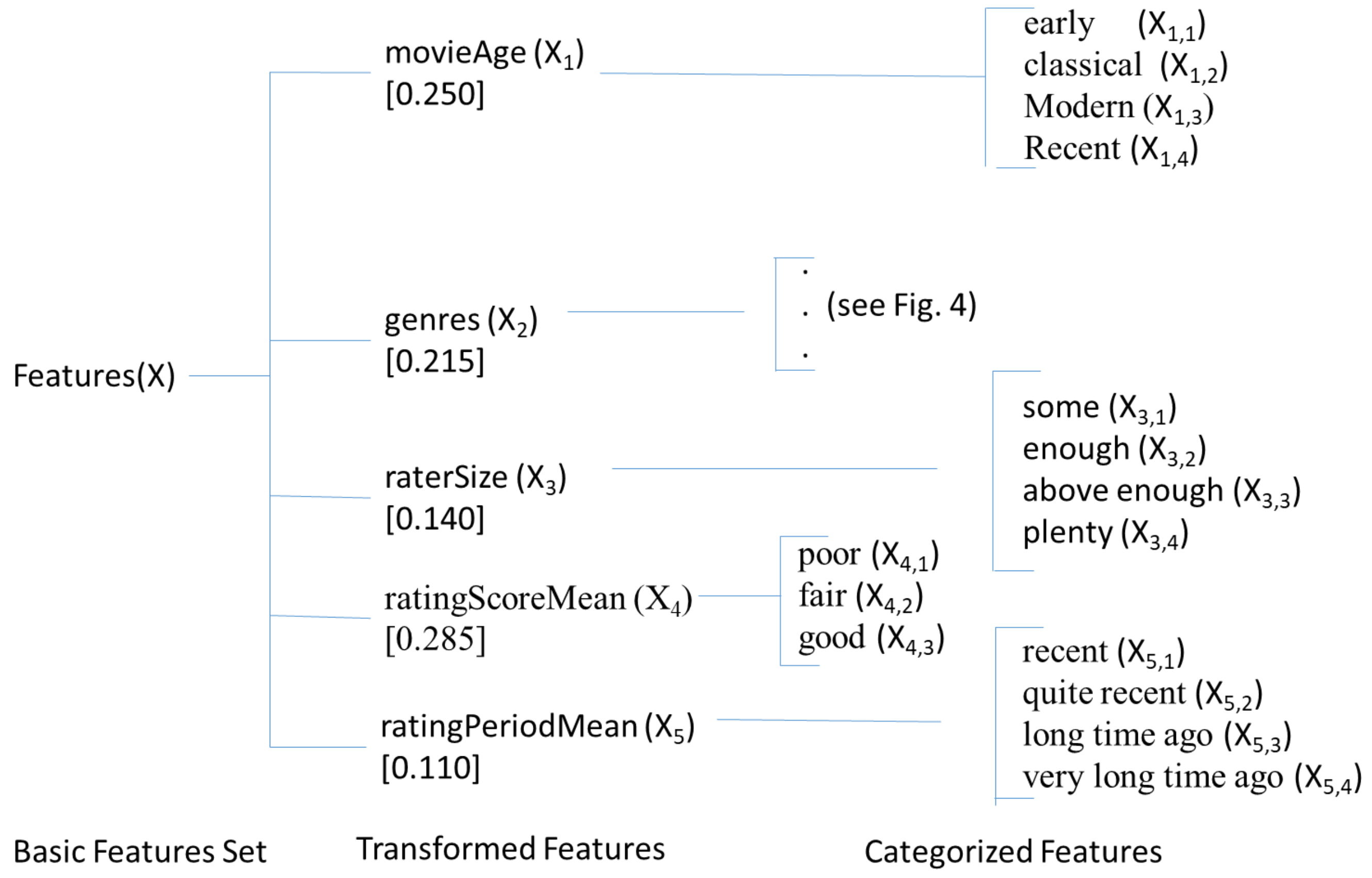

Figure 5: Transformed features with CPC features Categorization

Table 9: Transformed Features with CPC with respect to each machine learning category

<table>
<tr><th rowspan="2">Raw features</th><th colspan="4">Transformed features</th></tr>
<tr><th>Regression</th><th>Clustering</th><th>Classification</th><th>Association</th></tr>
<tr><td>movieId & title</td><td colspan="4">Identifier columns used for movie tracking; excluded from model training.</td></tr>
<tr><td>year</td><td colspan="3">$movieAge' = (X'_1) = 2019 - year$, where 2019 is the latest year of the dataset; As the newer movie is more preferred, the rescaling value is as below.<br>$movieAge = \nabla^-(movieAge') = X_1 = \nabla^-(X'_1)$</td><td>4 dummy variables via CPC binning: year: early, middle, modern</td></tr>
<tr><td>genres</td><td colspan="4">It is categorised as 18 dummy variables: Adventure, Animation, Children, Comedy, Fantasy, Romance, Drama, Action, Crime, Thriller, Horror, Mystery, Sci.Fi, Documentary, War, Musical, Western, Film.Noir. The weights are presented in Figure 4.</td></tr>
<tr><td>UserID</td><td colspan="3">raterSize ($X_3$): normalized number of unique users based on UserID to rate each movie<br>$raterSize = \nabla^+(raterSize') = X_3 = \nabla^-(X'_3)$</td><td>4 Dummy variables created from raterSize: some, enough, above enough, plenty;</td></tr>
<tr><td>rating</td><td>Aggregate the rating scores from different users for each movie, and store the results in the variable ratingScoreMean ($X'_5$), which is the target variable for regression.</td><td>ratingScoreMean ($X_5$) is rescaled as below.<br>$X_5 = \triangle^+(X'_5)$</td><td>Convert the ratingScoreMean variable as a categorical target variable called grade: poor, fair, good;</td><td>3 Dummy variables created from grade: poor, fair, good.</td></tr>
<tr><td>timestamp</td><td colspan="3">ratingYearMean: average of years that the raters rated.<br>ratingPeriodMean: ratingYearMean – year<br>$ratingPeriodMean = \nabla^+(ratingPeriodMean'_j) = X_4 = \nabla^+(X'_4)$</td><td>Dummy variables created from ratingPeriodMean: recent, quite recent, quite some long time ago, long time ago</td></tr>
</table>

Feature variables undergo domain-specific rescaling to harmonize feature ranges between 0 and 1 while preserving preference directions. For positive features where higher raw values indicate higher user preference (e.g., raterSize, $X_3$), the standard min-max scaling function $\nabla^+$ is applied.

$$X_j = \nabla^+(X'_j) = \frac{(X'_j - \min(X'_j))}{\max(X'_j) - \min(X'_j)} \tag{25}$$

When positive features possess a non-zero lower bound (e.g., ratingScoreMean, $X_5$ where minimum scores start at 0.5), the dividing maximal scaling function $\triangle^+$ is deployed to maintain proportional ratio relationships:

$$X_j = \triangle^+(X'_j) = \frac{X'_j}{\max(X'_j)} \tag{26}$$

Conversely, for negative features where lower raw values are preferred by users—specifically movieAge ($X_1$) (where newer films are favored) and ratingPeriodMean ($X_4$) (where recent rating activity is favored)—the max-min scaling function $\nabla^-$ inverts raw values into positive preference orientations:

$$X_j = \nabla^-(X'_j) = \frac{(\max(X'_j) - X'_j)}{\max(X'_j) - \min(X'_j)} \tag{27}$$

A complete mapping of raw MovieLens metadata attributes to transformed feature representations across machine learning task categories is summarized in Table 9.

### 6.3. Regression with CPC

A regression model is constructed to predict movie rating scores based on the feature structures shown in Figures 4-5, and Table 9. Linear regression is a well-established supervised learning method used to analyze relationships between predictor variables and a continuous target variable. Feature variable selection and representative dataset partitioning are critical prerequisites for ensuring model quality, stability, and predictive performance. Figure 6 summarizes the CPC-guided regression analysis workflow. Within this pipeline, experts utilize Cognitive Pairwise Comparisons as an interactive interface to define training-validation-test split ratios (Algorithm 1) and determine feature preference orderings (Algorithm 2) for constructing, validating, and evaluating the regression model. The technical procedures and computational modules for this section are hosted via the open-source web link: https://kkfyuen.github.io/cpcHCADDemos/s3cpcRegression.html. Details are outlined below.

$$y_i = \left(\beta_0 + \sum_{j=1}^{N} \beta_j x_{i,j}\right) + \varepsilon_i \tag{28}$$

$$y_i = \hat{y}_i + \varepsilon_i, \qquad i = 1, \dots, M \tag{29}$$

In matrix form,

$$Y = \beta \cdot X + \varepsilon = \hat{Y} + \varepsilon \tag{30}$$

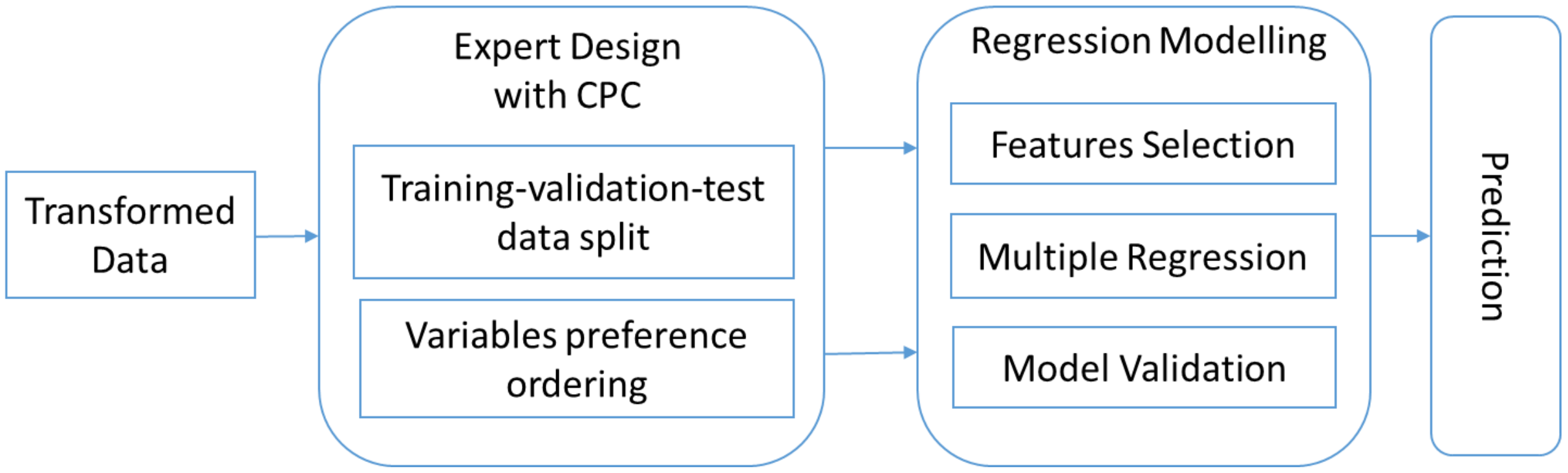

Figure 6: CPC for regression analysis

$X$ is a matrix of values containing the intercept column $X_0 = \{1\}_{(n+1)\times 1}$ and all predictor feature variables ($X_{j\neq 0}$). $\beta = \{\beta_0, \beta_1, \dots, \beta_N\}^T$ represents the vector of the intercept ($\beta_0$) and regression coefficients ($\{\beta_{j\neq 0}\}$). $\hat{y}_i$ denotes the predicted rating score corresponding to ground truth value $y_i$ . To estimate the optimal coefficient vector $\beta$, Ordinary Least Squares (OLS) minimizes the total sum of squared residual errors:

$$\begin{aligned} \text{Min}\left(\sum \epsilon_i^2\right) &= \text{Min}\left(\sum_{i=1}^{M} (y_i - \hat{y}_i)^2\right) \\ &= \text{Min}\left(\sum_{i=1}^{M} \left(y_i - \left(\beta_0 + \sum_{j=1}^{N} \beta_j x_{i,j}\right)\right)^2\right) \end{aligned} \tag{31}$$

The closed-form solution to this OLS optimization problem is expressed as:

$$\beta = \frac{Y}{X} = \frac{X^T Y}{X^T X} = (X^T X)^{-1} X^T Y \tag{32}$$

If $\det\left(X^T X\right) \neq 0$, $X^T X$ is invertible and yields a unique solution. In practice, QR decomposition is employed to perform matrix inversion reliably. Model goodness-of-fit across datasets is evaluated using the Mean Squared Error (MSE):

$$MSE(X, Y, \beta) = \frac{\sum_{i=1}^{M} \left(y_i - \left(\beta_0 + \sum_{j=1}^{N} \beta_j x_{i,j}\right)\right)^2}{M - N - 1} \tag{33}$$

Defining appropriate split ratios for training, validation, and testing subsets is inherently subjective. Cognitive Pairwise Comparisons resolve this by eliciting domain expert preferences for partitioning splits through $B_{tvt}$ ($\kappa_{tvt} = 4$, AI = 0.0963), deriving proportional split weights $W_{tvt} = \{0.778, 0.139, 0.083\}$). An empirical demonstration of dataset partitioning via Algorithm 1 on the sanitized MovieLens dataset (M = 54,266 movies) is detailed in Table 10. Specifically, the dataset is partitioned into a training set $\tilde{X}_{trn}$ ($L_1$ = 42,206 rows, 77.78%), a validation set $\tilde{X}_{val}$ ($L_2$ = 49,743 rows, 13.89%), and a testing set $\tilde{X}_{tst}$ (54,266 rows, 8.33%).

Table 10: Training-validation-test datasets splitting with CPC for a transformed and cleaned movie dataset

| $M = 54266$ , $\kappa_{tvt}$=4 , AI =0.0963 | | | | | | | |
|---|---|---|---|---|---|---|---|
| index | $B_{tvt}$ | trn | val | tst | $w_i$ | $L_t$ | $\tilde{X}_i$ |
| 1 | trn | 0 | 8 | 8 | 0.777778 | 42206 | $\tilde{X}_{(1,\ldots,42206)}$ |
| 2 | val | -8 | 0 | 1 | 0.138889 | 49743 | $\tilde{X}_{(42207,\ldots,49743)}$ |
| 3 | tst | -8 | -1 | 0 | 0.083333 | 54266 | $\tilde{X}_{(49744,\ldots,54266)}$ |

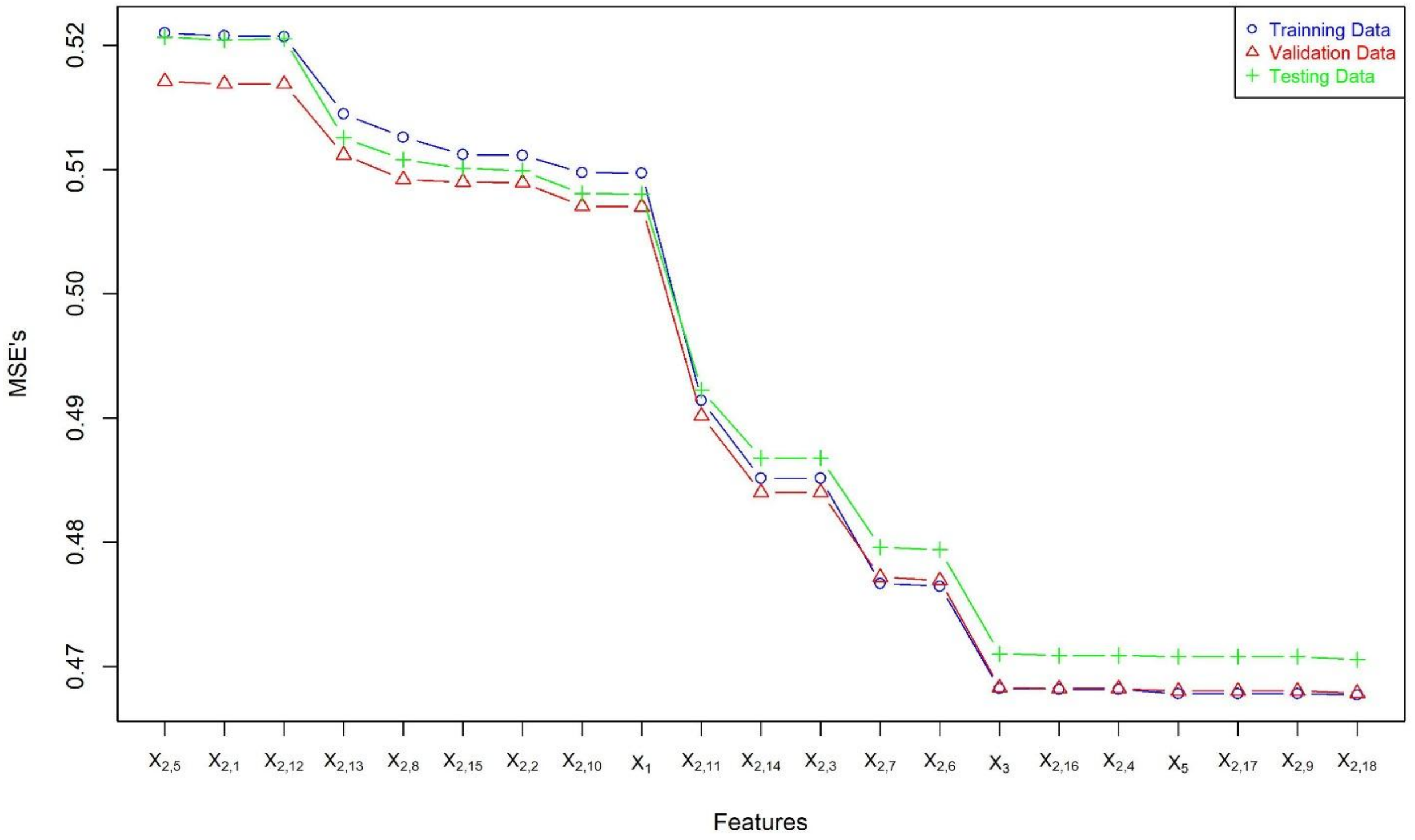

Figure 7: Mean square errors with forward features selection with training-validation-test split and CPC

Selecting an optimal minimal feature subset from $N$ candidate variables requires evaluating among $2^N$ possible feature combinations (including the intercept-only model). To systematically guide the search while incorporating domain expertise, the Forward Feature Selection with CPC algorithm (Algorithm 2) is deployed. Candidate features are evaluated sequentially according to their overall CPC priority weights ($W_{imp}$, computed by multiplying top-level category weight $W_{impF}$ by genre sub-level weights $W_{Genres}$). For instance, Fantasy ($X_{2,5}$) holds the highest initial priority (0.3404 = 0.215 × 1.5833), followed by Adventure ($X_{2,1}$, 0.3330), Mystery ($X_{2,12}$, 0.3270), Sci.Fi ($X_{2,13}$, 0.3150), Action ($X_{2,8}$, 0.3061), and War ($X_{2,15}$, 0.2897)). A candidate feature variable is accepted and permanently added to the active regression model only if its inclusion simultaneously reduces MSE on both training and validation sets:

- $\psi(\beta_{trn}^{+}) = -MSE(X_{trn}, Y_{trn}, \beta_{trn}^{+}) > \psi(\beta_{trn}) = -MSE(X_{trn}, Y_{trn}, \beta_{trn})$ and
- $\psi(\beta_{val}^{+}) = -MSE(X_{val}, Y_{val}, \beta_{val}^{+}) > \psi(\beta_{val}) = -MSE(X_{val}, Y_{val}, \beta_{val})$

As summarized in Table 11 and illustrated by the MSE trajectory plot in Figure 7, four candidate features—Children ($X_{2,3}$), Comedy ($X_{2,4}$), Western ($X_{2,17}$), and Crime ($X_{2,9}$)—failed to yield performance improvements on the validation set and were accordingly excluded from the final model. The 17 selected predictor features achieve highly consistent MSE values across all three dataset splits: $MSE_{trn}$= 0.46769, $MSE_{val}$= 0.46785, and $MSE_{tst}$ = 0.47055. Furthermore, 15 of the 17 selected predictor coefficients demonstrate strong statistical significance ($p < 0.01$), with *raterSize* ($X_3$, $\beta = +2.7241$, $p < 0.001$)), *movieAge* ($X_1$, $\beta = -0.6904$, $p < 0.001$)), and *ratingPeriodMean* ($X_5$, $\beta = -0.6684$, $p < 0.001$)) exhibiting the strongest predictive influence on movie rating scores ($R^2 = 0.1031$, $F = 203.7$, $p < 2.2 \times 10^{-16}$).

Table 11: Feature selection based on CPC priorities and mean of square errors with additional feature

| Notations | Names | Priorities | $MSE_{trn}$ | $MSE_{val}$ | $MSE_{tst}$ | Coefficients |
|---|---|---|---|---|---|---|
| $X_0$ | (intercept) | | 0.5211 | 0.5172 | 0.5208 | 3.6300*** |
| $X_{2,5}$ | Fantasy | 0.3404 | 0.5210 | 0.5171 | 0.5207 | 0.0346. |
| $X_{2,1}$ | Adventure | 0.333 | 0.5208 | 0.5169 | 0.5204 | -0.0286. |
| $X_{2,12}$ | Mystery | 0.327 | 0.5207 | 0.5169 | 0.5206 | 0.0587** |
| $X_{2,13}$ | Sci.Fi | 0.315 | 0.5145 | 0.5112 | 0.5126 | -0.1785*** |
| $X_{2,8}$ | Action | 0.3061 | 0.5126 | 0.5092 | 0.5108 | -0.0994*** |
| $X_{2,15}$ | War | 0.2897 | 0.5112 | 0.5090 | 0.5101 | 0.1305*** |
| $X_{2,2}$ | Animation | 0.2717 | 0.5112 | 0.5089 | 0.5099 | 0.1189*** |
| $X_{2,10}$ | Thriller | 0.2523 | 0.5098 | 0.5071 | 0.5081 | -0.0169 |
| $X_1$ | movieAge | 0.25 | 0.5097 | 0.5070 | 0.5081 | -0.6904*** |
| $X_{2,11}$ | Horror | 0.2255 | 0.4914 | 0.4902 | 0.4923 | -0.3287*** |
| $X_{2,14}$ | Documentary | 0.2031 | 0.4851 | 0.4840 | 0.4868 | 0.4181*** |
| $X_{2,3}$ | Children | 0.1822 | 0.4851 | 0.4840 | 0.4868 | ---- |
| $X_{2,7}$ | Drama | 0.1612 | 0.4767 | 0.4772 | 0.4796 | 0.2063*** |
| $X_{2,6}$ | Romance | 0.1403 | 0.4764 | 0.4769 | 0.4794 | 0.0384** |
| $X_3$ | raterSize | 0.14 | 0.4682 | 0.4683 | 0.4710 | 2.7241*** |
| $X_{2,16}$ | Musical | 0.1224 | 0.4682 | 0.4682 | 0.4709 | 0.0423 |
| $X_{2,4}$ | Comedy | 0.1135 | 0.4682 | 0.4682 | 0.4709 | ---- |
| $X_5$ | ratingPeriodMean | 0.11 | 0.4678 | 0.4680 | 0.4708 | -0.6684*** |
| $X_{2,17}$ | Western | 0.103 | 0.4678 | 0.4680 | 0.4708 | ---- |
| $X_{2,9}$ | Crime | 0.0941 | 0.4678 | 0.4680 | 0.4708 | ---- |
| $X_{2,18}$ | Film.Noir | 0.0896 | 0.4677 | 0.4678 | 0.4705 | 0.1413** |

*--Significant codes: 0 '***' 0.001 '**' 0.01 '*' 0.05 '.' 0.1 ' ' 1*

### 6.4. Classification with CPC

Evaluating and selecting the most appropriate classification algorithm for a specific domain task is a fundamental challenge in machine learning pipelines, as competing models exhibit disparate strengths across different evaluation metrics. In this application, Cognitive Pairwise Comparisons for model selection (Section 5.7, Eqs. 18–20) is deployed to evaluate and rank seven candidate classification algorithms across five multi-criteria performance metrics. A conceptual workflow summarizing the CPC classification pipeline is presented in Figure 8. The technical procedures and computational modules for this section are hosted via the open-source web link: https://kkfyuen.github.io/cpcHCADDemos/s4cpcClassification.html.

Predictor feature variables correspond to the 22-feature architecture (comprising 4 continuous/scaled attributes, i.e., *movieAge* ($X_1$), *raterSize* ($X_3$), *ratingPeriodMean* ($X_4$), and 18 CPC-weighted genre dummy variables [$X_{2,1}$ ,.... $X_{2,18}$]) extracted from 54,266 sanitized MovieLens records (Table 9). The target variable *Grade* represents the categorical movie quality, categorized into three ordinal classes—*poor*, *fair*, and *good*—using CPC discretization cut points (Table 6) derived from user rating score averages (*ratingScoreMean*).

To establish a rigorous benchmarking suite, seven widely adopted classification algorithms implemented in open-source *R* packages are benchmarked: Deep Learning in Keras (Alliare & François, 2026), Single-layer Neural Network (Venables & Ripley, 2002), C5.0 Decision Tree (Kuhn & Quinlan, 2025), k-Nearest Neighbors with k = 5 (Venables & Ripley, 2002), Multinomial Logistic Regression (Venables & Ripley, 2002), Support Vector Machine with Radial Basis Function (Meyer et al., 2025), and Naïve Bayes (Meyer et al., 2025). All models utilize standard or default package hyperparameter configurations to facilitate reproducible research.

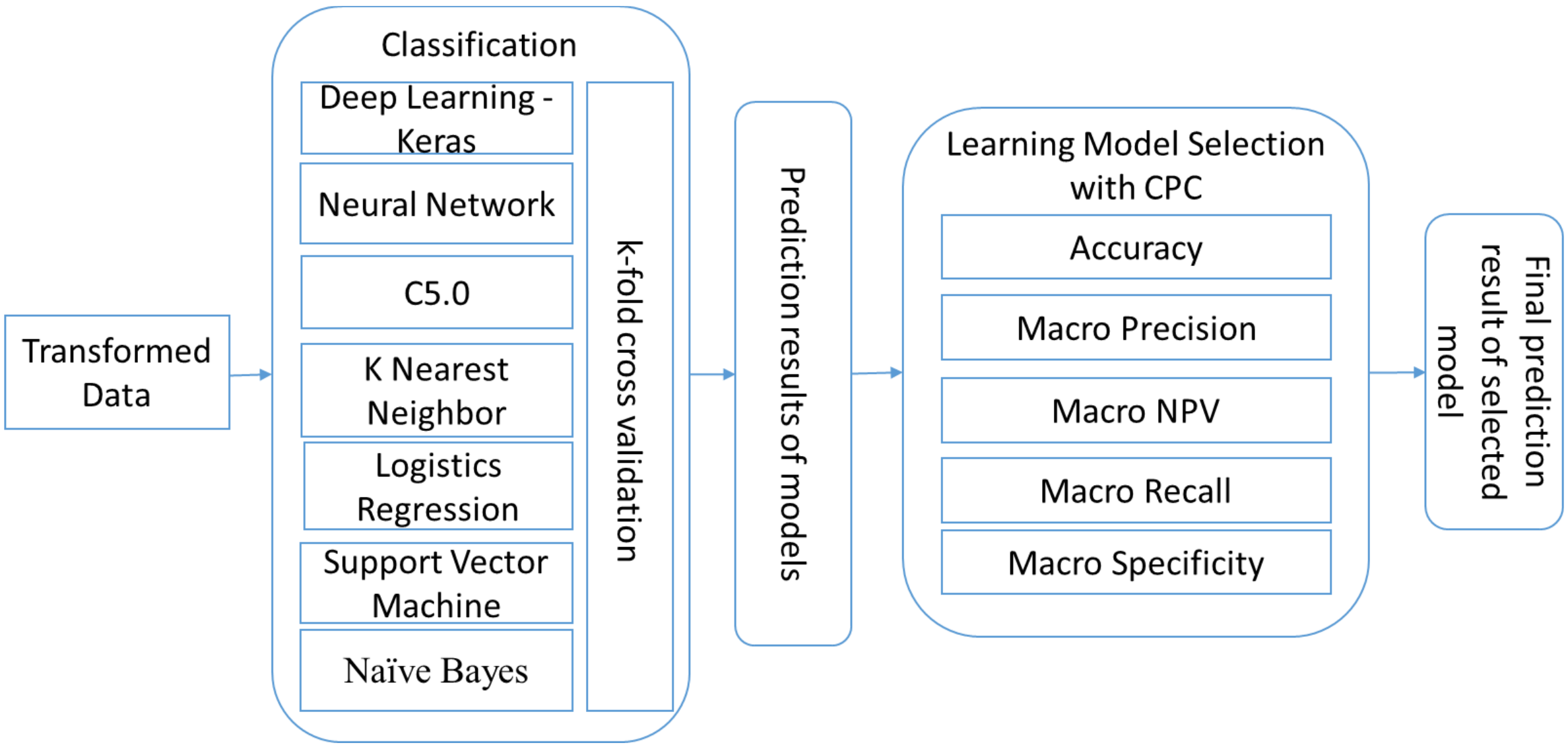

Figure 8: CPC for classification analysis

Table 12: notations and classification benchmark functions

| | |
|---|---|
| $TP_j$: true positive count for class label *j;*<br>$TN_j$: true negative count for class label *j;*<br>$FP_j$: false positive count for class label *j;*<br>$FN_j$: false negative count for class label *j;*<br>L: number of class labels. | $Accuracy = \frac{TP + TN}{TP + TN + FP + FN}$ |
| $macroNPV = \frac{1}{L}\sum_j \frac{TN_j}{TN_j + FN_j}$ | $macroPrecision = \frac{1}{L}\sum_j \frac{TP_j}{TP_j + FP_j}$ |
| $macroRecall = \frac{1}{L}\sum_j \frac{TP_j}{TP_j + FN_j}$ | $macroSpecificity = \frac{1}{L}\sum_j \frac{TN_j}{TN_j + FP_j}$ |

Classifier performance is evaluated across five multi-class benchmarking functions defined in Table 12: Accuracy, Macro Precision, Macro Negative Predictive Value (NPV), Macro Recall, and Macro Specificity. These functions aggregate binary confusion matrix elements—True Positives ($TP_j$), True Negatives ($TN_j$), False Positives ($FP_j$), and False Negatives ($FN_j$)—across all *L=3* target class labels (*poor*, *fair*, and *good*).

Table 13: Criteria weights for classification benchmarks with CPC

| POM | Accuracy | Macro Precision | Macro NPV | Macro Recall | Macro Specificity | Weight |
|---|---|---|---|---|---|---|
| Acc. | 0 | -2 | 3 | -2 | 3 | 0.210 |
| Macro Precision | 2 | 0 | 5 | 0 | 5 | 0.260 |
| Macro NPV | -3 | -5 | 0 | -5 | 0 | 0.135 |
| Macro Recall | 2 | 0 | 5 | 0 | 5 | 0.260 |
| Macro Specificity | -3 | -5 | 0 | -5 | 0 | 0.135 |

To diversify evaluation techniques within the framework, *10-fold cross-validation* is utilized instead of the training-validation-test split strategy deployed in regression. Because standard cross-validation partitions datasets into fixed, equal folds, CPC is not applied to fold generation. Observations are shuffled using a fixed seed (seed = 1000) and partitioned into 10 equal folds. Model performance is computed across all 10 folds; for example, the 10-fold cross-validation accuracy vector for SVM is (0.750, 0.756, 0.765, 0.755, 0.742, 0.752, 0.762, 0.765, 0.760, 0.764), yielding a mean accuracy of 0.757, which is shown in Table 14.

Subjective decision-maker preferences across the five benchmarking metrics are elicited via a Pairwise Opposite Matrix ($POM_{cri}$) with root value $\kappa_{cri}$ = 8, as detailed in Table 13. The resulting matrix is perfectly accordant (*Accordance Index AI = 0*), and processing via the RAU operator yields the criteria weight vector: $W_{cri}$ = {Accuracy: 0.210, Macro Precision: 0.260, Macro NPV: 0.135, Macro Recall: 0.260, Macro Specificity: 0.135}. Precision and Recall receive the highest preference weights (0.260 each), reflecting decision-makers' prioritization of positive class predictive accuracy.

By combining the 10-fold cross-validation performance matrix with the CPC criteria weights (Table 13), the CPC-weighted decision matrix for learning model selection is constructed (Section 5.7, Eqs. 18–20) and presented in Table 14. The weighted aggregation function (Eq. 20) yields overall decision scores for ranking candidate classifiers.

The decision scores establish Support Vector Machine (SVM) as the optimal classification model with the highest composite decision score ($z$ = 0.622, closely followed by C5.0 Decision Tree ($z$ = 0.620, Rank 2) and Multinomial Logistic Regression ($z$ = 0.615, Rank 3). SVM achieves the highest Macro Precision (0.676) and highly robust Accuracy (0.757) and Macro NPV (0.784). Although C5.0 achieves slightly higher Accuracy (0.760) and Macro Recall (0.387), SVM's superior precision performance under the CPC-derived preference weights elevates it to Rank 1. Conversely, untuned Deep Learning (Keras, z = 0.300, Rank 7) and Neural Network (z = 0.328, Rank 6) perform poorly under default settings, highlighting the necessity of human preference-weighted multi-criteria evaluation rather than relying solely on single-metric accuracy.

Table 14: CPC weighted decision matrix for learning models selection

| W | Acc. 0.210 | Macro Precision 0.260 | Macro NPV 0.135 | Macro Recall 0.260 | Macro Specificity 0.135 | Score | Rank |
|---|---|---|---|---|---|---|---|
| keras | 0.213 | 0.084 | 0.419 | 0.334 | 0.667 | 0.300 | 7 |
| nnet | 0.147 | 0.175 | 0.582 | 0.322 | 0.662 | 0.328 | 6 |
| c5.0 | 0.760 | 0.615 | 0.786 | 0.387 | 0.696 | 0.620 | 2 |
| knn | 0.729 | 0.47 | 0.732 | 0.395 | 0.7 | 0.571 | 4 |
| logReg | 0.757 | 0.647 | 0.788 | 0.348 | 0.674 | 0.615 | 3 |
| svm | 0.757 | 0.676 | 0.784 | 0.346 | 0.673 | 0.622 | 1* |
| Naïve Bayes | 0.387 | 0.402 | 0.7 | 0.441 | 0.707 | 0.490 | 5 |

### 6.5. User Preference Weighted Data Reduction for Clustering with CPC

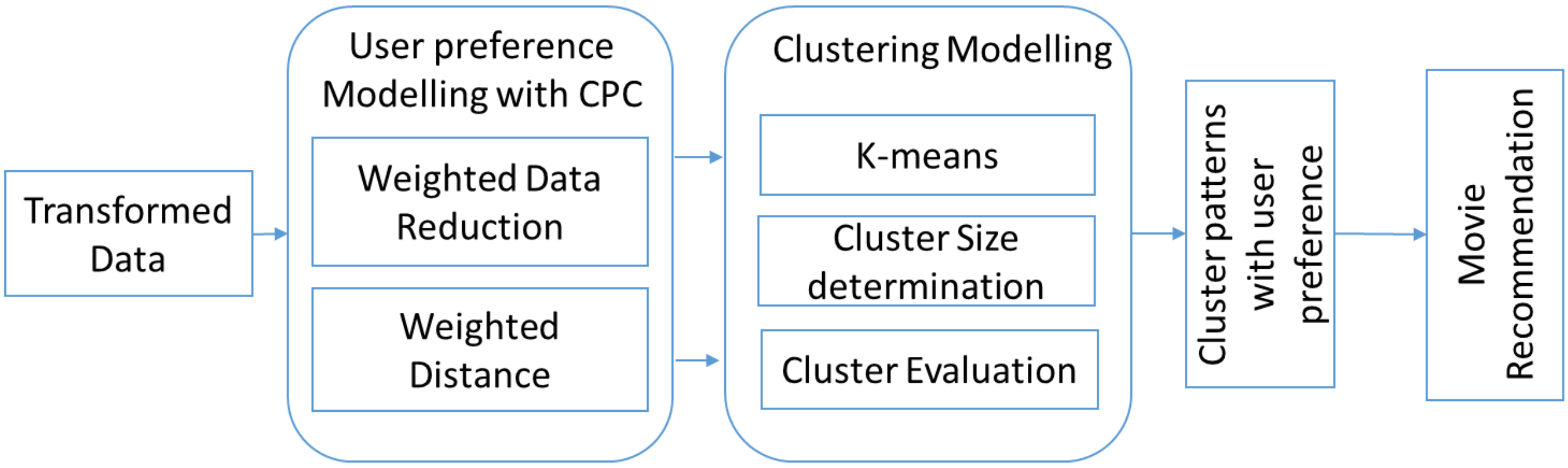

Figure 9: User preference involved Data Reduction for Clustering with CPC

Figure 9 presents a conceptual overview of user-preference-guided data reduction and clustering within the Cognitive Pairwise Comparisons for Human-Centric Artificial Intelligence and Data Science (CPC-HCAD) framework. Principal Component Analysis (PCA) is a widely established data reduction technique, whilst K-Means clustering is a popular unsupervised learning method used to uncover latent patterns in multidimensional data. The integration of PCA and K-Means guided by CPC resolves a key limitation of conventional data reduction—namely, that standard PCA operates blindly to domain priorities. The technical procedures and computational modules for this section are hosted via the open-source web link: https://kkfyuen.github.io/cpcHCADDemos/s5cpcClustering.html.

In this application, CPC models subjective evaluator preferences across movie features (Figure 4) to construct weighted principal components (Algorithm 3) and weighted Euclidean distance metrics (Theorem 1, Eqs. 21–24). The input dataset comprises 22 rescaled features (comprising 4 continuous/scaled attributes—movieAge ($X_1$), raterSize ($X_3$), ratingPeriodMean ($X_4$), and ratingScoreMean ($X_5$)—alongside 18 CPC-weighted genre dummy variables ($X_{2,1}$ ,..., $X_{2,18}$)) across 54,266 sanitized MovieLens records (Table 9). Feature variables are rescaled into non-negative [0, 1] preference ranges using $\nabla^+(X_j)$ min-max scaling rather than z-score standardization, ensuring non-negative feature representations suitable for preference aggregation.

Figure 10 illustrates the scree plot of eigenvalues alongside the cumulative variance proportion curve for the 22 user-preference-weighted principal components. Because standard PCA is sensitive to feature variances, assigning higher CPC weights amplifies target feature variances, ensuring that top principal components reflect user domain priorities. Applying an eigenvalue threshold of eigenvalue >= 0.9 selects the first 14 principal components (comp1 through comp14). These 14 components capture 80.0% (0.800) of total cumulative variance, effectively reducing feature dimensionality by 8 dimensions across 54,266 observation records (saving 434,128 values from downstream clustering computations) while preserving essential preference signals.

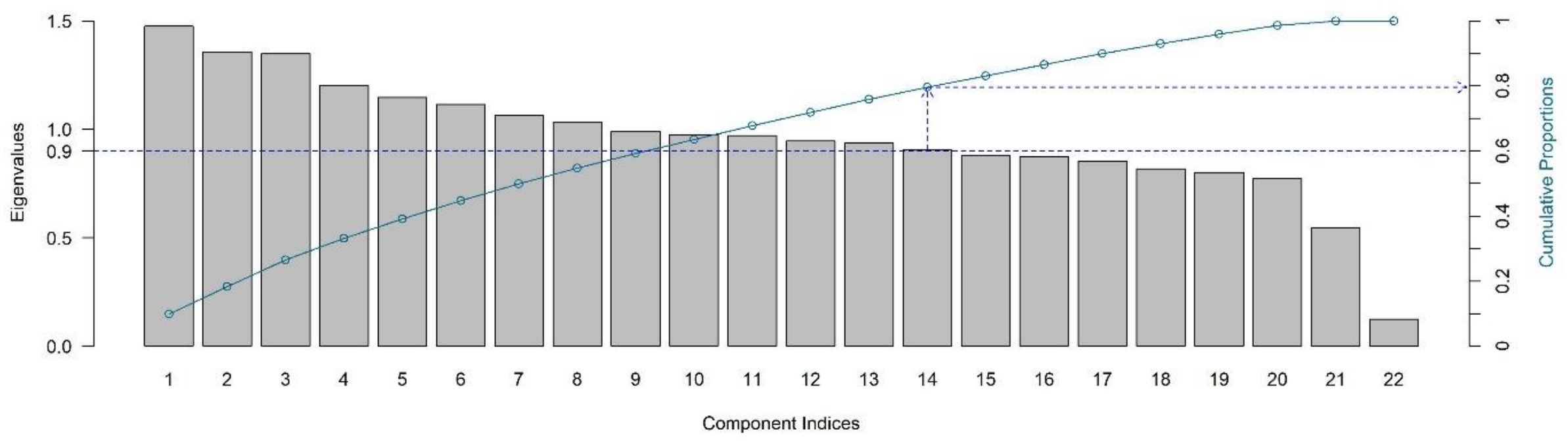

Figure 10: Selection of principal components

Following preference-weighted data reduction, K-means clustering is executed on the 14 selected weighted principal components using MacQueen's algorithm (MacQueen, 1967) as outlined below.

1. Select $K$ data points as initial cluster centroids (μ).
2. Compute the Euclidean distance from each data point ($x_i$) to all $K$ centroids.
3. Assign each data point ($x_i$) to its nearest cluster ($I_k$).
4. Update the centroid positions (μ) for all $K$ clusters.
5. Repeat Steps 2–4 until cluster assignments stabilize.

The algorithm is implemented via the *kmeans* function in R (R Core Team, 2026), setting the hyperparameter configuration to *iter.max* = 200 and *nstart* = 100. The objective of the clustering process is to minimize the Total Within-Cluster Sum of Squares ($TWCSS$), defined as follows

$$TWCSS = \sum_{k=1}^{K} \sum_{i \in I_k} \sum_{j=1}^{n} \left(x_{ij} - \mu_{kj}\right)^2 \quad (34)$$

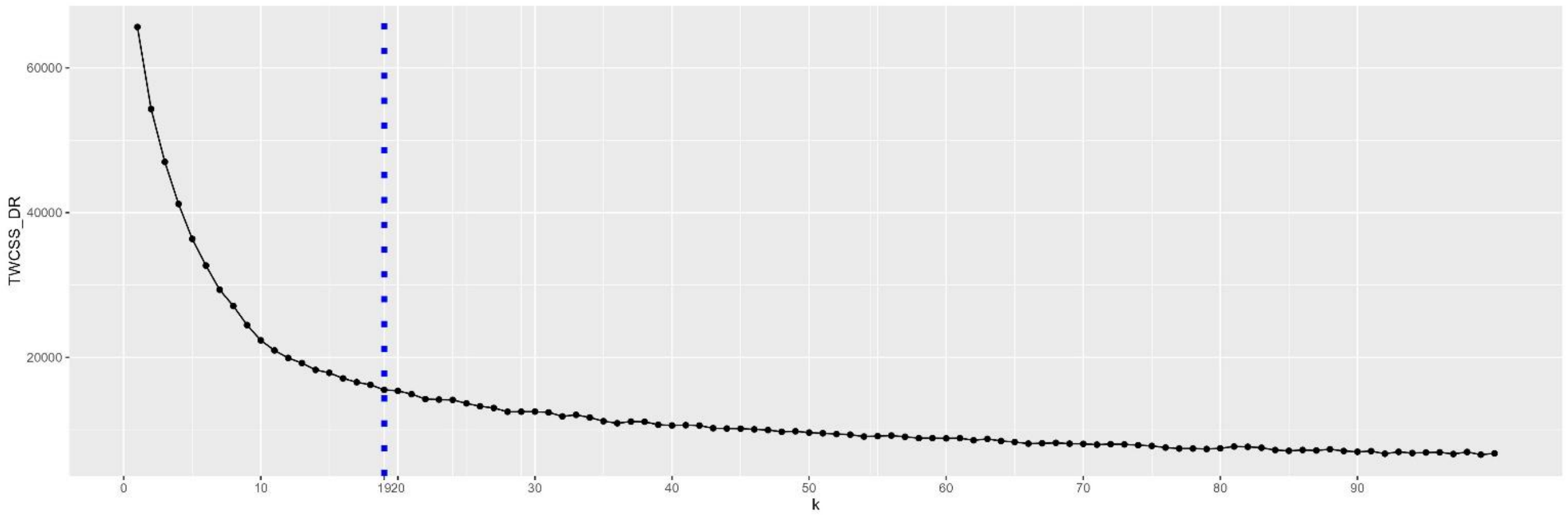

Figure 11: Determination of optimal cluster number by elbow/knee point

To determine the optimal number of clusters (k), TWCSS is evaluated across k = 1 to 100. Because K-Means initialization is sensitive to random seed placement, 100 random restarts are executed for each k, retaining the solution with minimal TWCSS. As shown in Figure 11, the Python kneed package (Satopaa et al, 2011) is used in *R* to detects the convex elbow/knee point on the TWCSS trajectory curve at exactly $k = 19$ clusters. At $k = 19$, the Proportion of Variance Explained (PVE, computed as Between-SS / Total-SS ratio) achieves 76.2% (0.762), confirming a highly distinct, well-separated cluster structure.

Theorem 1 shows Weighted Euclidean Distance of both $X$ and $W$ is the same as the Euclidean Distance of $\dot{X} = \sqrt{W}X$, i.e., $wed(W, X) = ed(\dot{X})$. To incorporate the weights into the data, $\dot{X}$ is used for the function of the Average Silhouette Value (ASV) as below.

$$\text{ASV} = \frac{1}{K}\sum_{k=1}^{K}\left(\frac{1}{|I_k|}\sum_{i \in I_k}\frac{b_i - a_i}{max(a_i, b_i)}\right) \tag{35}$$

$$a_i = \frac{1}{|I_k| - 1}\sum_{\grave{i} \in I_k, i \neq \grave{i}} ed(\dot{x}_i, \dot{x}_{\grave{i}}) = \frac{1}{|I_k| - 1}\sum_{\grave{i} \in I_k, i \neq \grave{i}} (\dot{x}_i - \dot{x}_{\grave{i}})^2 \tag{36}$$

$$b_i = \min_{k \neq i}\frac{1}{|I_k|}\sum_{\grave{i} \in I_k} ed(\dot{x}_i, \dot{x}_{\grave{i}}) = \min_{k \neq i}\frac{1}{|I_k|}\sum_{\grave{i} \in I_k}(\dot{x}_i - \dot{x}_{\grave{i}})^2 \tag{37}$$

The ASV metric ranges between -1 and 1. $a_i$ denotes the mean distance between a data point $\dot{x}_i$ and all other data points $\{\dot{x}_{\grave{i} \neq i} : \grave{i}, i \in I_k\}$ in the same cluster $k$, whilst $b_i$ represents the minimum average distance from $\dot{x}_i$ in cluster $i$ to all points in neighbouring clusters i.e., $\{\dot{x}_{\grave{i}} : \grave{i} \in I_{k \neq i}\}$. Evaluated on the 14 selected principal components, the clustering model achieves an ASV of 0.314; evaluated on squared preference-weighted raw data, the model achieves an ASV of 0.251. Both validation scores confirm acceptable cluster cohesion and separation.

Table 15: Top-ranked 2019 movie recommendations from the cluster containing Toy Story (1995)

| Preference | Id | title | genres | Year | $X'_5$ |
|---|---|---|---|---|---|
| 0.721 | 208939 | Klaus | Adventure\|Animation\|Children\|Comedy | 2019 | 4.313 |
| 0.706 | 205066 | Nezha | Animation\|Comedy\|Fantasy | 2019 | 3.750 |
| 0.700 | 208241 | White Snake | Animation\|Fantasy\|Romance | 2019 | 3.500 |
| 0.682 | 201588 | Toy Story 4 | Adventure\|Animation\|Children\|Comedy | 2019 | 3.624 |
| 0.670 | 207828 | Birthday Wonderland | Animation\|Fantasy | 2019 | 2.000 |
| 0.669 | 206714 | Abominable | Adventure\|Animation\|Children\|Comedy | 2019 | 3.406 |
| 0.667 | 203222 | The Lion King | Adventure\|Animation\|Children\|Drama | 2019 | 3.149 |
| 0.664 | 207347 | Spread Your Wings | Adventure\|Children | 2019 | 2.750 |
| 0.661 | 200630 | Missing Link | Adventure\|Animation\|Children\|Comedy\|Fantasy | 2019 | 2.857 |

To generate targeted movie recommendations, movies across the dataset are ranked by user preference values calculated as weighted aggregates using top-level and genre sub-level CPC weights (Figure 4, Table 9). Across all 54,266 titles, the movie with the highest overall preference score is Toy Story (1995, preference = 0.786). The cluster containing Toy Story (1995) comprises 905 titles with an average cluster preference score of 0.6637. Filtering this cluster for recent titles (2019) yields a targeted recommendation list sharing the preference profile of Toy Story (1995). Table 15 displays these top-ranked recent movie recommendations extracted from the cluster, led by Klaus (preference = 0.721, ratingScoreMean = 4.313), Nezha (0.706, 3.750), White Snake (0.700, 3.500), Toy Story 4 (0.682, 3.624), Birthday Wonderland (0.670, 2.000), and Abominable (0.669, 3.406).

### 6.6. Associate rule analysis with CPC

Association rule learning (or association rule mining) discovers underlying co-occurrence patterns and decision rules to predict the likelihood of specific item occurrences based on antecedent items in transaction databases. Figure 12 illustrates the conceptual workflow of Cognitive Pairwise Comparisons for association rule analysis in a human-centric movie recommender system. Because conventional association rule algorithms (such as Apriori) process discrete categorical inputs exclusively, continuous numerical feature variables cannot be directly utilized. Within the proposed CPC-HCAD framework, continuous variables are discretized into structured categorical intervals using CPC preference-proportional binning (Eqs. 12–14), which are subsequently transformed into binary dummy variables via one-hot encoding (Figure 5, Table 9). Once feature data formats satisfy algorithmic input requirements, candidate rules generated by association mining are evaluated and ranked according to user preference weights, ensuring that recommendations reflect human domain priorities. The technical procedures and computational modules for this section are hosted via the open-source web link: https://kkfyuen.github.io/cpcHCADDemos/s6cpcAssociation.html.

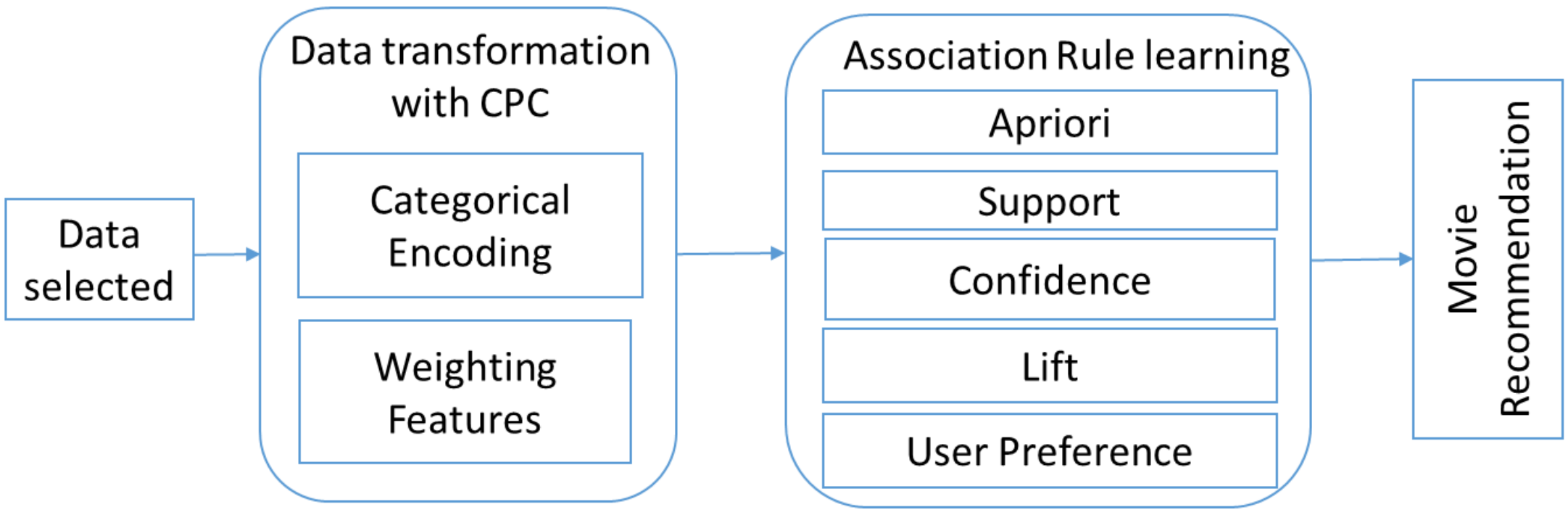

Figure 12: CPC for Associate rule analysis for movie recommendation

To establish itemset representations for transaction mining, let the set of all candidate movies be defined as a set of items $I = \{I_1, \dots, I_i, \dots, I_m\}$, and let the set of feature attributes correspond to a transaction matrix $T = \{t_1, \dots, t_j, \dots, t_n\}$. Because individual movies share common attribute categories, each transaction $t_j$ contains a subset of items termed an *itemset*. A $k$-itemset denotes a collection containing exactly $k$ items, while an empty set containing no items represents the null itemset. Fundamental concepts and formal properties of association rule algorithms are thoroughly reviewed in Tan (2019).

Table 16: Discretisation with CPC for movie years

| | sf = 1.0075 | | min(Year) = 1874 | | max(Year) = 2019 | AI = 0.08 | | |
|---|---|---|---|---|---|---|---|---|
| index | | Early | Classical | Modern | Recent | W | h | H |
| 1 | Early | 0 | 4 | 6 | 8 | 0.391 | 1944.7 | [-∞,1944.7) |
| 2 | Classical | -4 | 0 | 3 | 6 | 0.289 | 1986.6 | [1044.7,1986.6) |
| 3 | Modern | -6 | -3 | 0 | 3 | 0.203 | 2016.1 | [1986.6, 2016.1) |
| 4 | Recent | -8 | -6 | -3 | 0 | 0.117 | | [2016.1,∞] |

Feature selection and feature transformation for association rule learning are detailed in Figure 5 and Table 9. Continuous attributes—specifically *year*, *raterSize*, *ratingScoreMean*, and *ratingYearMean* —are discretized via CPC preference-proportional binning into 33 binary dummy variables (4 + 18 + 4 + 3 + 4 = 33). Table 6 and Table 16 detail computational discretization steps for movie rating scores and release years, respectively. As shown in Table 16, movie release years (ranging from 1874 to 2019) are partitioned into four historical periods—Early, Classical, Modern, and Recent—using a soft factor sf = 1.0075 and Accordance Index AI = 0.080, yielding cut points at 1944.7, 1986.6, and 2016.1.

The Apriori algorithm implemented in the open-source *R* arules package (Hahsler et al., 2026) is deployed for association rule learning. An association rule consists of two disjoint itemsets: an antecedent $A$ and a consequent $B$, expressing an *If A, then B* predictive relationship, as defined as below.

$$A \to B, where\ A, B \in I, A \cap B = \emptyset \quad (38)$$

The mining process comprises two main computational phases: frequent itemset generation governed by a minimal support threshold, followed by rule derivation governed by a minimal confidence threshold. Core mathematical metrics for association rule learning are summarized in Table 17.

Table 17: notations and functions for associate rule learning

| Count: $count(A) = \{t_i : A \subseteq t_i \in T\}$ | Support: $support(A \to B) = \frac{count(A \cup B)}{N}$ |
|---|---|
| Confidence: $conf(A \to B) = \frac{count(A \cup B)}{count(A)}$ | Lift: $lift(A \to B) = \frac{support(A \to B)}{support(A) \times support(B)}$ |

In this application, the minimal support parameter is set to supp = 2/33 ≈ 0.061, while the minimal confidence parameter is set to conf = 0.90. Across the transformed dataset, the number of candidate items is m = 54,338 movies, evaluated across n = 33 feature transaction columns. The maximal itemset length $|A \cup B| = 6$ specifies a 5-item antecedent $A$ predicting a single target movie consequent $B$ in each generated rule. Table 18 displays an exemplary watch history comprising five recent films evaluated by a user. Executing the Apriori algorithm generates 1,902 association rules matching this antecedent. To deliver tailored recommendations, candidate consequents are filtered for recent releases (release year ≥ 2018) and ordered strictly by user preference values derived as weighted feature aggregates (Figure 4, Table 9). Table 19 presents the top-10 recommended movies delivered by the system.

Table 18: An example of a watched list of recent five movies for a user

| movieId | title | genres | year | *RSM* | Preference |
|---|---|---|---|---|---|
| 4993 | Lord of the Rings: The Fellowship of the Ring | Adventure\|Fantasy | 2001 | 4.091 | 0.892 |
| 5952 | Lord of the Rings: The Two Towers | Adventure\|Fantasy | 2002 | 4.068 | 0.884 |
| 194288 | The Monkey King 3 | Action\|Adventure\|Fantasy | 2018 | 5.000 | 0.860 |
| 8368 | Harry Potter and the Prisoner of Azkaban | Adventure\|Fantasy | 2004 | 3.811 | 0.823 |
| 72998 | Avatar | Action\|Adventure\|Sci-Fi | 2009 | 3.622 | 0.804 |

Table 19: Top *N* recommendation where *N*=10

| movieId | title | genres | year | *RSM* | Sup. | Conf. | Lift | Preference |
|---|---|---|---|---|---|---|---|---|
| 195159 | Spider-Man: Into the Spider-Verse | Action\|Adventure\|Animation\|Sci-Fi | 2018 | 4.111 | 0.061 | 1.000 | 3.667 | 0.795 |
| 208293 | The Seven Deadly Sins: Prisoners of the Sky | Action\|Adventure\|Animation\|Fantasy | 2018 | 4.000 | 0.061 | 1.000 | 3.000 | 0.790 |
| 201773 | Spider-Man: Far from Home | Action\|Adventure\|Sci-Fi | 2019 | 3.713 | 0.061 | 1.000 | 3.667 | 0.782 |
| 201026 | Mia and the White Lion | Adventure | 2018 | 3.500 | 0.061 | 1.000 | 3.667 | 0.782 |
| 203322 | The Peanut Butter Falcon | Adventure | 2019 | 3.393 | 0.061 | 1.000 | 3.300 | 0.776 |
| 194823 | Genghis Khan | Action\|Adventure\|Fantasy | 2018 | 3.333 | 0.061 | 1.000 | 3.667 | 0.765 |
| 182715 | Annihilation | Adventure\|Mystery\|Sci-Fi\|Thriller | 2018 | 3.565 | 0.061 | 1.000 | 4.125 | 0.762 |
| 189333 | Mission: Impossible - Fallout | Action\|Adventure\|Thriller | 2018 | 3.746 | 0.061 | 1.000 | 3.667 | 0.762 |
| 186607 | The Pagan King | Action\|Adventure | 2018 | 3.357 | 0.061 | 1.000 | 4.125 | 0.760 |
| 201488 | Modest Heroes | Adventure\|Animation\|Fantasy | 2018 | 3.400 | 0.061 | 1.000 | 3.667 | 0.758 |

As demonstrated in Table 19, combining association rule mining with Cognitive Pairwise Comparisons overcomes a key limitation of standard transaction mining—namely, that unweighted rule outputs order candidate consequents strictly by support or confidence metrics without incorporating user domain preferences. By re-ranking candidate rule consequents using CPC-derived preference values, the system prioritizes highly rated, preference-aligned cinematic titles (e.g., Spider-Man: Into the Spider-Verse*, preference = 0.795*) that directly match the user's implicit genre and temporal priorities.

## 7. Conclusions and Discussion

This study successfully advances and extends Cognitive Pairwise Comparisons (CPC) to directly program human domain knowledge and subjective priorities into AI&DS algorithmic pipelines. Unlike prior decision-making studies that treat CPC strictly as an isolated multi-criteria evaluation tool, this research introduces novel programmatic formulations (including Algorithms 1–3) that adapt CPC into low-level data preprocessing, feature selection, and machine learning operations. Within the proposed framework, CPC operates as an interactive, user-facing interface that empowers data scientists, domain experts, and end users to shape machine learning workflows according to explicit preference constraints, thereby mitigating automation bias and capturing tacit domain knowledge.

While existing literature in Human-Centric AI and Data Science (HCAD) predominantly focuses on qualitative design guidelines, conceptual surveys, or post-hoc interpretability tools, it lacks an interval-scale mathematical mechanism capable of translating subjective human judgments into programmatic pipeline operations. The proposed framework directly bridges this operational gap. Specifically, this study makes three core technical contributions to the HCAD literature:

1. Programmatic CPC Pipeline Extensions: Novel extensions of the CPC method were established to address critical preference-elicitation challenges across seven foundational data science pipeline stages:
    i. Categorical encoding via interval preference matrices, replacing arbitrary one-hot or ordinal assignments (Eqs. 9–11);
    ii. Continuous data discretization using preference-proportional binning (Eqs. 12–14);
    iii. Training-validation-test data partitioning based on cognitive prioritization (Algorithm 1);
    iv. Human-guided forward feature selection (Algorithm 2);
    v. Multi-criteria learning model selection (Eqs. 18–20);
    vi. User-preference-weighted Principal Component Analysis (Algorithm 3); and
    vii. Weighted Euclidean distance metric integration (Theorem 1, Eqs. 21–24) for clustering and association rule learning.
2. Cross-Paradigm Algorithmic Integration: The proposed human-centered formulations were seamlessly integrated into established machine learning paradigms, encompassing linear regression, multi-class classification, weighted Principal Component Analysis, K-Means clustering, and Apriori association rule mining.
3. End-to-End Recommender System Validation: The practical applicability, validity, and robustness of the CPC-HCAD framework were empirically validated using the open-source MovieLens 25M dataset (54,266 sanitized records), demonstrating how human-in-the-loop interactions yield highly personalized, reliable, and preference-aligned cinematic recommendations.

Despite its demonstrated efficacy in embedding tacit human judgment into machine learning pipelines, the proposed CPC-HCAD framework has several inherent limitations:

1. Cognitive Load and Pairwise Scalability: Complete pairwise comparison requires $n(n-1)/2$ evaluations. While hierarchical feature decomposition and criteria grouping effectively mitigate evaluator fatigue for moderate feature spaces, eliciting preferences for ultra-high-dimensional datasets (e.g., thousands of unstructured attributes) remains cognitively demanding for human experts.
2. Dependence on Evaluator Expertise: The quality of the derived pipeline weights depends directly on the domain knowledge and consistency of the human evaluator. Although the Accordance Index ($AI \leq 0.1$) provides a strict mathematical gatekeeper for matrix consistency, biased or erroneous human judgment can still propagate into model parameters.
3. Static Preference Elicitation: Evaluator preferences are currently captured upfront via Pairwise Opposite Matrices (POM) as static weight configurations, rather than being continuously updated through real-time, interactive feedback loops.

To address these limitations, future research will focus on: (i) developing sparse, incomplete pairwise comparison models to reduce evaluator fatigue in high-dimensional feature spaces; (ii) extending the CPC-HCAD framework to dynamic, real-time reinforcement learning settings where human preference weights update adaptively during system usage; and (iii) scaling the methodology to emerging state-of-the-art architectures, including deep neural networks, transformer-based language models, and multi-modal industrial applications.

**Acknowledgements:** The author is very grateful to the Editor-in-Chief, Associate Editor, and anonymous referees for their time and valuable efforts to improve and recommend this work.

**Funding:** This research was not funded by any company, external grant, or commercial organization.

**Author Contribution:** The entire work, including theoretical formulation, algorithmic design, software implementation, and empirical validation, was conducted solely by K. K. F. Yuen.

**Conflicts of Interest / Competing Interests:** The author declares that there are no financial, personal, or professional conflicts of interest regarding the publication of this paper.

**Consent for Publication:** The present work did not involve clinical trials, physical human experiments, or private human subject data.

**Data Availability Statement:**

- The raw datasets used in this study are openly available in the MovieLens collection at https://grouplens.org/datasets/movielens/25m/ (MovieLens, 2019).
- All processed datasets, computational web modules for core algorithms, and supplementary materials are accessible via the author's GitHub repository at https://github.com/kkfyuen/cpcHCADDemos and permanently archived on Zenodo at https://doi.org/10.5281/zenodo.23434993 (Yuen, 2026).
- Live computational web module demos are hosted on GitHub Pages at https://kkfyuen.github.io/cpcHCADDemos/ (Yuen, 2026).